\documentclass[preprint]{elsarticle}

\usepackage{array}
\usepackage{amsfonts}       
\usepackage{amsmath}
\usepackage{amssymb}
\usepackage{booktabs} 
\usepackage{subcaption,float}
\usepackage{xcolor}
\usepackage[colorlinks,hypertexnames=true]{hyperref}       
\usepackage{url}            
\usepackage{microtype} 
\usepackage{array}
\usepackage{graphicx}
\usepackage[colorinlistoftodos]{todonotes}
\usepackage[linesnumbered,ruled,vlined,boxed]{algorithm2e}
\usepackage[misc]{ifsym}
\usepackage{import}
\usepackage{lineno}

\journal{Journal of \LaTeX\ Templates}

\newcommand{\vf}{{\boldsymbol{f}}}
\newcommand{\vx}{{\mathbf{x}}}
\newcommand{\vtau}{{\boldsymbol{\tau}}}
\newcommand{\vy}{{\mathbf{y}}}
\newcommand{\vu}{{\mathbf{u}}}
\newcommand{\vz}{{\mathbf{z}}}
\newcommand{\vzero}{{\mathbf{0}}}
\newcommand{\vs}{{\mathbf{s}}}
\newcommand{\bbR}{{\mathbb{R}}}

\newcommand{\bbV}{{\mathbb{V}}}
\newcommand{\cov}{{\mathrm{Cov}}}
\newcommand{\gp}{{\mathcal{GP}}}
\newcommand{\compO}{{\mathcal{O}}}
\newcommand{\Low}{{\mathcal{L}}}
\newcommand{\N}{{\mathcal{N}}}
\newcommand{\F}{{\mathcal{F}}}
\newcommand{\D}{{\mathcal{D}}}

\newcommand{\loss}{{\mathcal{L}}}
\newcommand{\vk}{{\mathbf{k}}}
\newtheorem{remark}{Remark}
\newcommand{\tbf}[1]{\textbf{#1}}

\newcommand{\eref}[1]{Eq.~\eqref{eq:#1}}
\newcommand{\fref}[1]{Fig.~\ref{fig:#1}}
\newcommand{\sref}[1]{Section~\ref{sec:#1}}
\newcommand{\tref}[1]{Table~\ref{tab:#1}}
\newcommand{\cref}[1]{Chapter~\ref{chp:#1}}
\newcommand{\aref}[1]{\algo~\ref{algo:#1}}

\newcommand{\algo}{\tbf{Algorithm}}
\newcommand{\freq}[1]{\hat{#1}} 
\newcommand{\timedelay}{{\theta}}
\newcommand{\phasedelay}{{\phi}}
\newcommand{\invF}{{{\F}_{s\rightarrow \tau}^{-1}}}
\newcommand{\Four}{{{\F}_{\tau\rightarrow {s}}}}

\newcommand{\Var}{{\Sigma}}

\newcommand{\tra}{^{\top}}
\newcommand{\vmu}{{\boldsymbol{\mu}}}
\newcommand{\vtime}{{\boldsymbol{\timedelay}}}
\newcommand{\stime}{{{\timedelay}}}
\newcommand{\vphase}{{\boldsymbol{\phasedelay}}}

\newcommand{\sphase}{{{\phasedelay}}}

\newcommand{\uptext}{\overbrace}
\newcommand{\upline}{\overline}

\newcommand{\gcsm}{\text{SMD}}
\newcommand{\sm}{\text{SM}}
\newcommand{\itimej}{{{i}\times{j}}}
\newcommand{\gcsmij}{_{{\gcsm}}^{\itimej}}

\newcommand{\rest}{\text{rest}}
\newcommand{\init}{\text{init}}

\newcommand{\nlml}{\text{NLML}}
\newcommand{\prune}{\text{prune}}

\newcommand{\sa}{\text{SA}}
\newcommand{\compressr}{\text{CR}}
\newcommand{\sr}{\text{SR}}

\newcommand{\best}{\text{best}}

\newcommand{\se}{\text{SE}}

\newcommand{\gibbs}{\text{Gibbs}}
\newcommand{\std}[1]{{\scalebox{0.8}{\hspace{0.3mm}$\pm$\hspace{0.3mm}#1}}}

\begin{document}
    
\title{Compressible Spectral Mixture Kernels with Sparse Dependency Structures for Gaussian Processes}

\begin{frontmatter}
    
    \author{Kai Chen$^{a}$, Feng Yin$^{b}$(\Letter),              
        and Shuguang Cui$^{b, c}$
    }
    
    \address[a]{School of Mathematics and Statistics, Central South University, Changsha 410083, China}
    \address[b]{School of Science and Engineering (SSE), The Chinese University of Hong Kong, Shenzhen 518172, China}
    \address[c]{Future Network of Intelligence Institute (FNii), The Chinese University of Hong Kong, Shenzhen 518172, China}
    
    \begin{abstract}%
        Spectral mixture (SM) kernels comprise a powerful class of generalized kernels for Gaussian processes (GPs) to describe complex patterns. This paper introduces model compression and time- and phase (TP) modulated dependency structures to the original (SM) kernel for improved generalization of GPs. Specifically, by adopting Bienaym\'es identity, we generalize the dependency structure through cross-covariance between the SM components. Then, we propose a novel SM kernel with a dependency structure (SMD) by using cross-convolution between the SM components. Furthermore, we ameliorate the expressiveness of the dependency structure by parameterizing it with time and phase delays. The dependency structure has clear interpretations in terms of spectral density, covariance behavior, and sampling path. To enrich the SMD with effective hyperparameter initialization, compressible SM kernel components, and sparse dependency structures, we introduce a novel structure adaptation (SA) algorithm in the end. A thorough comparative analysis of the SMD on both synthetic and real-life applications corroborates its efficacy.
    \end{abstract}
    
    
\end{frontmatter}

\section{Introduction}\label{sec:intro}
{G}{}aussian processes (GPs) constitute an important class of Bayesian nonparametric models for machine learning \cite{Theodoridis2020}. A GP models an underlying system by applying a Gaussian prior to the underlying function and computes the posterior distribution over this function given the observations. 
This allows GPs to learn a function approximation well if a sufficient number of observations is accumulated. Furthermore, GPs can avoid overfitting in cases where only a little evidence \cite{Rasmussen2010,Rasmussen2006} is available. A GP can model a large class of systems by selecting and designing the kernel function, which reflects the autocovariance structure of the system. However, similar to other kernel learning methods, such as support vector machines (SVMs), selecting kernel function is one of the most important factors for GP model because an expressive kernel determines the representation ability of GP, and the posterior distribution can change significantly by using different kernels. 

For GPs, however, a kernel is usually selected subjectively, heavily depending on expert knowledge and empirical analysis of data patterns. There are a handful of base kernels and their combinations for diversified GP leaning applications, such as received signal strength (RSS)-based radio map modeling \cite{yin2017distributed} and wireless traffic prediction \cite{Xu19,Yin2020FedLoc,chen2022recent}. 
To avoid human intervention, automatic \cite{Wilson2013,Duvenaud2013,yin2020linear} and generalized kernel designs \cite{Dai2020interpretable} are highly demanded for GPs. 
Since \cite{Wilson2013} introduced an automatic and expressive kernel, called spectral mixture (SM) kernel, by modeling the spectral density (SD) of a stationary signal with a sum of Gaussians. 
GPs with the SM kernel
have been successfully employed in various applications, such as medical time series prediction \cite{Duerichen2015a}, Arctic coastal erosion forecasting \cite{Kupilik2017}, and urban environmental monitoring using sensor networks \cite{chen2019multioutput}.

Briefly, there are many advances focusing on diversified extensions of SM kernel rather than the latent dependency structure between SM components.
The SM product (SMP) kernel \cite{Wilson2014a} constructed via a product of several SM kernels on the input dimensions can model image and spatial data.
The non-stationary SM (NSM) kernel \cite{Remes2017,Herlands2016} demonstrates a compelling ability to represent input-dependent covariances between inputs. 
The grid SM (GSM) kernel \cite{yin2020linear} linearly combines independent low-rank sub-kernels to approximate the underlying covariance of a stationary signal.
However, all these variants of SM kernel cannot explicitly represent the dependency structures. 

In this paper, we show that there are extensive dependency structures between the SM components. 
We propose a new SM kernel with a dependency structure (SMD).
In addition, we introduce a structure adaptation (SA) algorithm to compress the SMD effectively. We demonstrate the benefits of modeling the dependency structure and applying model compression.

More specifically, we generalize the dependency structure as a cross covariance between the SM components by using Bienaym\'e's identity for the linear superposition form of GP. The 
cross covariance
is constructed by a cross convolution between the SM components. 
To ameliorate the expressiveness of the dependency structure,
we design a complex-valued Gaussian SD that incorporates both time and phase (TP) delays for SM component. 
We then construct a positive definite SMD kernel that handles the dependency structures.
The spectral density, covariance behavior, and sampling path of the dependency structure are interpretable and informative. 
We propose a SA algorithm, including bootstrap-based hyperparameter initialization (BHI), compression of SMD components, and sparsification of the dependency structures, to improve the learning efficiency and interpretability of the SMD automatically.
The SA algorithm can prevent the hyperparameter space from expansion and retain valuable dependency structures of SMD.
To analyze the dependency structure,
we introduce a measure of dependency intensity $\gamma_{ij}$ (see \eref{learned-corr}).
The SMD kernel can be regarded as the generalization of the original SM kernel; that is, by only considering the autocovariance of its components, the SMD kernel will reduce to the SM kernel. 
Preliminary results of this work have been presented in \cite{kai2019}. While this paper includes abundant new contributions, such as:
\begin{itemize}
    \item We represent the dependency structure as a generalized cross covariance.
    \item We construct a complex-valued Gaussian mixture model (GMM) characterized by TP delays to model the SDs. 
    \item A new SMD kernel 
    demonstrates good interpretability, and more importantly better expressiveness.
    \item 
    An effective and interpretable SA algorithm 
    for compressible SMD. 
    \item We investigate the interpolation and extrapolation performances, scalable learning, dependency structures, compression ratio (CR), 
    and sparsity ratio (SR), 
    of the SMD on multiple synthetic and real-life datasets.
\end{itemize}

The rest of this paper is organized as follows. Background on GPs and SM kernels and a summary of the related works are given in \sref{back}.
In \sref{motivation}, we present our motivation.
\sref{gcsm} introduces our SMD kernel.
\sref{hyp} describes the SA algorithm for the SMD. 
\sref{exp} shows multiple experiments.
Concluding remarks and future works are discussed in \sref{conclude}.


\section{Background and related work}\label{sec:back}
In this section, we review GPs, SM kernels, and related works.

\subsection{Gaussian processes}\label{sec:gp}

Given an observation pair $\{\vx_{i} \in{\bbR}^{P}, y_{i}\}$ with $P$ dimensional input,  a GP can represent a function map $y=f(\vx)+\epsilon$ to approximate the underlying system, where $\epsilon$ is the noise. In this paper, we mainly consider 
real-valued GPs. 
From the function-space view, a GP \cite{Theodoridis2020,Rasmussen2006} defines a prior distribution 
over functions, completely specified by its first-order and second-order statistics, namely, the mean function $m({\vx})$ and the covariance function $k({\vx}, {{\vx}{'}})$. 
In general, a GP is defined 
as $f({\vx})\sim{\gp}(m({\vx}), k({\vx}, {{\vx}{'}}))$.
We use the terms covariance function, kernel, and kernel function interchangeably.

Given a selected kernel function $k({\vx}, {{\vx}{'}})$ and training set $\D=\{X, \vy\}$ for a GP model, we can analytically compute the predictive mean $\tilde{\vy}^{*}$ and variances $\bbV[{\vy}^{*}]$ (that is, its predictive uncertainty) for the testing set ${X}^*$ by using the inferred posterior distribution $p(\vy^{*}|X^{*}, X, \vy)\sim\N(\tilde{\vy}^{*}, \bbV[{{\vy}^{*}}])$.
$k({\vx}, {{\vx}{'}})$ has free hyperparameters $\Theta$ determining the model complexity of a GP.
A GP model is optimized by minimizing the negative log-marginal likelihood (NLML), $\loss_{\nlml}\triangleq-\log\ p({\vy}|{X},{\Theta})$, 
where $\loss$ is obtained through marginalization over the latent function $f(X)$ \cite{Theodoridis2020,Rasmussen2006}.


\subsection{Spectral mixture kernels}\label{sec:sm}
The SM kernel \cite{Wilson2013} has been derived with the aid of Bochner's Theorem \cite{Bochner2016,stein2012interpolation}.
The essence of this theorem is that 
a function $k$ on $\bbR^P$ is the covariance function of a weakly stationary mean square continuous complex-valued random process on $\bbR^P$ if and only if it can be represented as $k(\vtau) = \int_{\bbR^P} e^{2\pi \imath\vs^{\top}\vtau}d\psi(\vs)$, 
where $\psi$ is a positive finite measure and  $\imath$ denotes the imaginary unit.
If $\psi$ has density $\freq{k}({\vs})$, then $\freq{k}$ is called the SD or power spectrum of the kernel. Moreover, $k$ and $\freq{k}$ constitute an Fourier transform (FT) pair; that is,
$\freq{k}(\vs)=\Four[k(\vtau)](\vs)$ and $k(\vtau)=\invF[\freq{k}(\vs)](\vtau)$, where the operators $\Four$ and $\invF$ denote the FT and the inverse FT, respectively. 
Originally, the SM kernel \cite{Wilson2013} was constructed by approximating the underlying SD as a mixture of $Q$ Gaussians in the frequency domain. The SM kernel can approximate any stationary kernel with a sufficient number of Gaussian components. 
By applying the inverse FT, we can obtain the SM kernel
as $k_{\sm}=\sum_{i=1}^Q  w_{i}k_{{\sm},i}(\vtau)$ with
\begin{align}\label{eq:smp}
    \begin{split}        
        k_{{\sm},i}(\vtau)
        &=\exp\big(-2\pi^2\vtau^{\top}{\Var}_{i}\vtau\big)\cos\big(2\pi\vtau\tra\vmu_{i}\big),
    \end{split}
\end{align}
where
$k_{{\sm},i}(\vtau)=\invF[\freq{k}_{{\sm},i}(\vs)](\vtau)$ is the $i$-th SM component, 
$w_i$, $\vmu_{i}
$, and ${\Var}_{i}
$ are  the signal magnitude, center frequency, and frequency bandwidth parameters of $k_{{\sm},i}(\vtau)$, respectively. 


\subsection{Related work}\label{sec:related}
There is rich literature on GPs related to SM kernel function design and analysis \cite{Rasmussen2006,Wilson2013,Duvenaud2013,Duvenaud2014,Wilson2014}. This section mainly focuses on the family of SM kernels \cite{Wilson2013,Wilson2014a,yin2020linear} and some new variants. More related SM extensions are surveyed in \cite{Remes2017,kai2019,yin2020linear}. Similar to the compositional form of the SM kernel, additive GPs \cite{Duvenaud2013,Duvenaud2014} implicitly sum over some 
one-dimensional base kernels to construct a flexible kernel representation. For the advances of SM kernel \cite{Wilson2014a,chen2021gaussian}, we have the SMP kernel with  $k_\text{SMP}(\vtau|\Theta)=\sum_{i=1}^{Q}\prod_{p=1}^{P}k_\sm(\vtau_{p}|\Theta_{p})$
and NSM kernel \cite{Remes2017,Herlands2016}. The NSM kernel includes
    a non-stationary Gibbs kernel $k_{\gibbs,i}(x,x')$ replacing the exponential part of the SM kernel, and input-dependent $w_i(x)$ and $\mu_i(x)$ corresponding to $w_i$ and $\mu_i$ in the SM kernel. Recently, an approach was proposed in \cite{kai2019} that encodes simple dependency structures between components of an SM kernel. 
    However, similar to the existing additive GPs, most of the existing SM variants assume that 
    GP components specified by SM components are independent and ignore their possible dependency structure.
    
    On the other hand, a quantification of the dependency structure between components in GPs was initially proposed in \cite{Duvenaud2014}. However, no further investigation in modeling the dependency structure is presented therein. In \cite{kai2019}, the main challenges in sparsifying the dependency structure and modeling TP delays of dependency structure
    are still unsolved. 

    In \cite{yin2020linear}, another hyperparameter efficient inference approach fixes $\vmu_{i}$ and $\Sigma_{i}$ of SM and allows only $w_{i}$ to be optimizable, which could result in a sparse SM kernel. In \cite{Jang2017}, a L\'evy process was introduced to automatically select the number of SM components, but the selection is not stable and easily encounters overfitting. 
    In short, the inference and optimization of the SMD in terms of hyperparameter initialization, model compression, and dependency structure sparsity have not yet been studied. 
    
    
    \section{Motivation}\label{sec:motivation}
    
    This section aims to give a generalized SM kernel with dependency structure and further comment on its properties. 
    All components are additive for the original SM kernel \cite{Wilson2013}. Any function $f$ drawn from a GP with the SM kernel $k_{\sm}$, that is, $f \sim \gp(0,k_{\sm})$, can be described as $f = \sum_{i=1}^Q f_{\sm, i}$,      
    where $f_{\sm, i} \sim \gp(0,w_i k_{{\sm},i})$.
    To simplify our notations, we use ${\vf}_{\sm, i}$ and ${\vf}_{\sm, i}^{*}$ to denote the respective function values evaluated on
    $X$ and 
    $X^*$, respectively. 
    
    \tbf{Generalization of dependency structure:} 
    Generally, by using the Bienaym\'e's identity \cite{klenke2006,loeve2017probability} for the linear form of $f$,
    the generalized covariance of $f$ is given by
    
    \begin{align}\label{eq:cov-full}
        \begin{split}
            \bbV[f]=&\sum_{i=1}^{Q}\sum_{j=1}^{Q}\cov(f_{\sm, i}, f_{\sm, j})\\
            =&\uptext{\sum_{i=j}\bbV[f_{\sm, i}]}^\text{autocovariance}+\uptext{\sum_{i\neq{j}}\cov(f_{\sm, i}, f_{\sm, j})}^\text{cross covariance},
        \end{split}
    \end{align}
    where $\bbV[f_{\sm, i}]=\cov(f_{\sm, i}, f_{\sm, i})$  
    is the autocovariance. 
    Here, the autocovariance of random function $f_{\sm, i}$ is computed as     
    $\cov(f_{\sm, i}, f_{\sm, i})=w_{i}k_{\sm, i}$.
    Therefore, we reformulate \eref{cov-full} as
    
    \begin{align}\label{eq:cov-dep}
        \bbV[f]=\sum_{i=1}^{Q}w_{i}k_{\sm, i}+\uptext{\sum_{i=1}^{Q}\sum_{j\neq{i}}^{Q}\cov(f_{\sm, i}, f_{\sm, j})}^\text{dependency structure}.
    \end{align}
    For the SM kernel, however, it restricts that $f_{\sm, i}$ and $f_{\sm, j}$ are  independent with 
    $\cov(f_{\sm, i}, f_{\sm, j})=0, \forall(i\neq{j})$. Unfortunately, there is no evident 
    support that $\cov(f_{\sm, i}, f_{\sm, j})=0$. 
    In this paper, we consider $\cov(f_{\sm, i}, f_{\sm, j})\neq{0}$ as a general dependency structure
    to free the SM kernel.

    
    \section{Spectral mixture kernel with dependency structure}\label{sec:gcsm}
    In this section, we propose for the first time an extended SM kernel incorporating dependency structure and its TP augmentations.
    
    \subsection{Modeling dependency structure using convolution}\label{sec:dependency}
    
    A stationary covariance function $k({\vx}, {\vx}')$ can be represented in convolution form on ${\bbR}^P$, as in \cite{Gaspari1999,Genton2015}, $k({\vx}, {\vx}')\ \triangleq\int_{{\bbR}^P} g({\vu})\ g({\vtau}-{\vu})\,{\rm{d}}{\vu}=(g*g)({\vtau})$, 
    where $\vtau \triangleq {\vx} - {\vx}'$ and $*$ denotes the convolution operator.
    Since convolution in the time domain corresponds to multiplication in the frequency domain,
    we have the squared form of the $i$-th SM component as 
    \begin{align}\label{eq:SM_half}
        \begin{split}
            {w_{i}\freq{k}_{{\sm},i}({\vs})}&=\Four[({g}_{{\sm},i}*{g}_{{\sm},i})({\vtau})](\vs)\\
            &=\freq{g}_{{\sm},i}^{2}({\vs}),
        \end{split}
    \end{align}
    where $\freq{g}_{{\sm},i}({\vs})$ is the SD of the $i$-th SM basis component. For the dependency structures $\cov(f_{\sm, i}, f_{\sm, j})$, one possible approach is to employ the cross correlation between functions  $f_{\sm, i}$ and $f_{\sm, j}$,  which is equal to the convolution of the two (weighted) kernels $w_{i}k_{{\sm},i}$ and $w_{j}k_{{\sm},j}$, namely,
    \begin{align}\label{eq:SM_cross}
        \begin{split}
            f_{\sm, i}\star f_{\sm, j}
            =\invF\left[w_{i}{\varphi}_{{\sm},i}({\vs})\cdot\upline{w_{j}{\varphi}_{{\sm},j}}({\vs}) \right](\vtau),
        \end{split}
    \end{align}
    where ${\varphi}_{{\sm},i}({\vs})=\N(\vs;{\vmu}_{i}, {\Var}_{i})$, $\invF$, $\star$, and $\upline{(-)}$ denote the inverse FT, the cross-correlation operator, and the complex conjugate operator, respectively. However, by directly using \eref{SM_cross}, we will obtain a kernel that is the inverse FT of the squared Gaussian $w_i^2\varphi^2_{{\sm},i}({\vs})$ when $i=j$. This is different from the original SM component.
    To ensure the compatibility between the dependency structure and SM component when $i=j$, 
    we, therefore, consider the convolution between the {\it basis components} $g_{\sm, i}(\vtau)$ and $g_{\sm, j}(\vtau)$, $g_{\sm, i}(\vtau)*g_{\sm, j}(\vtau)$, where $g_{\sm, i}(\vtau)=\invF[\freq{g}_{\sm, i}(\vs)](\vtau)$. Thus, we can describe $\cov(f_{\sm, i}, f_{\sm, j})$ well. Note that $\cov(f_{\sm, i}, f_{\sm, j})$ does not introduce additional parameters for the dependency structure.
    
    
    \subsection{Time and phase characterized Gaussian spectral density}\label{sec:complex-spec}
    
    In signal processing, for a signal sample of an underlying process, time delay differences between different signal frequency components characterize their temporal relationship and influence the signal's shape.
    Furthermore, due to the nature of FT, a signal in the frequency domain always has a complex representation with magnitude (real) and phase (imaginary) parts. 
    It can be of interest to know not only the magnitude but also the phase of the SD. The phase difference is useful for understanding the interference (dependency structure) phenomenon between components of a physical process. 
    However, the dependency structure investigated in \cite{kai2019} only paints a picture of the magnitude difference between the SM components. 
    Here, we introduce TP parameterization for the SD of the SM component to enrich its representation capacity. 
    
    Based on the property of FT, shifting a signal $k(\vtau)$ with time delay $\vtime$ in the time domain is equivalent to 
    multiplying a complex exponential in the frequency domain i.e., $\freq{k}_{\vtime}(\vs) = e^{-2\pi{\vtime}\vs{\,{\imath}}}\freq{k}(\vs)$, where $k_{\vtime}(\vtau) \triangleq k(\vtau-\vtime)$ and $\freq{k}(\vs)=\Four[k(\vtau)](\vs)$
    \cite{Bateman1954}.
    For any phase delay function $k_{\vphase}(\vtau)$ with a phase delay vector ${{\vphase}}$, the FT of $k_{\vphase}(\vtau)$ in the frequency domain is 
    $\freq{k}_{\vphase}(\vs) = e^{-{2\pi}{{\vphase}}{\,{\imath}}}\freq{k}(\vs)$.
    For the SMD kernel, we can directly embed the time delay $\vtime_{i}$ and phase delay $\vphase_{i}$ into the SD, $\freq{k}_{{\sm}, i}({\vs})$, yielding the following complex-valued {\it TP delay SD} function:
    \begin{align}\label{eq:gcsm_spec}
        \begin{split}
            \freq{k}_{{\gcsm}, i}({\vs})=&w_{i}\varphi_{{\sm}, i}({\vs})\uptext{\exp({-2\pi{\imath}\left({\vtime}_{i}{\vs}+{{\vphase}_{i}}\right)})}^{\text{TP delays}}.
        \end{split}
    \end{align}
    
    \subsection{Time- and phase modulated dependency structure}\label{sec:conv}
    Considering the TP modulated SD function in \eref{gcsm_spec} and adopting the squared form in \eref{SM_half}, 
    we define $ \freq{g}_{{\gcsm},i}({\vs}) =  \freq{k}_{{\gcsm},i}^{1/2}(\vs)$. 
    We then
    express the corresponding SD with the dependency structure  as 
    \begin{align}\label{eq:cross_gcsm_spec}
        \begin{split}
            \freq{k}\gcsmij({\vs})
            =&\freq{g}_{{\gcsm},i}({\vs})\cdot\upline{\freq{g}_{{\gcsm},j}}({\vs})\\
            =&{w_{ij}{a_{ij}}}{\varphi}_{{\gcsm},ij}({\vs})
            \uptext{\exp\big(-\pi{\imath}({\vtime}_{ij}{\vs}+{{\vphase}_{ij}})\big)}^{\text{cross TP delays}},
        \end{split}
    \end{align}
    with the following parameters:
    \begin{itemize}
        \item cross weight: 
        {$w_{ij}=\sqrt{w_{i}w_{j}}$},
        \item cross amplitude:
        ${a}_{ij}={\left|4\pi^{2}\Sigma_{i}\Sigma_{j}\right|}^{\frac{1}{4}}\N(\vmu_{i};\,\vmu_{j}, \frac{\Sigma_{i}+\Sigma_{j}}{2})$,
        \item cross Gaussian: $\varphi_{{\gcsm},ij}({\vs})=\N\big(\vs;\,\vmu_{ij}, \Sigma_{ij}\big)$
        \item cross mean of $\varphi_{{\gcsm},ij}$: 
        {$\vmu_{ij}=\frac {{\Var}_{i}{\vmu}_{j}+{\Var}_{j}{\vmu}_{i}}{{{\Var}_{i}+{\Var}_{j}}}$},
        \item cross covariance of $\varphi_{{\gcsm},ij}$: {${{\Var}}_{ij}=\frac{{2{{\Var}_{i}{\Var}_{j}}}}{{{\Var}_{i}+{\Var}_{j}}}$},
        \item cross time delay: {${\vtime}_{ij}={\vtime}_{i}-{\vtime}_{j}$}, 	
        \item cross phase delay: {${\vphase}_{ij}={\vphase}_{i}-{\vphase}_{j}$}. 	
    \end{itemize}

    The cross amplitude $a_{ij}$ is a normalization constant that only depends on the difference between components $i$ and $j$.
    Without TP delays, 
    the SD term in \eref{cross_gcsm_spec} can be reduced as follows: ${\hat{k}_{\gcsm, \vtime=0, \vphase=0}^{{i}\times{j}}(\vs)}\triangleq{\hat{g}_{{\sm},i}(\vs)}\cdot\upline{\hat{g}_{{\sm},j}}(\vs)$ \cite{kai2019}.

    
    \begin{remark} 
        According to \eref{cross_gcsm_spec}, 
        the closer the components are, the larger the weight $w_{ij}$, frequency $\vmu_{ij}$, and scale $\Sigma_{ij}$, and the greater 
        the dependency structure in the SMD. 
    \end{remark}
    
    \subsection{Spectral mixture with TP modulated dependency structure}\label{sec:tpsmk}
    In light of the SM kernel, by applying the inverse FT, we can define the dependency structure as 
    \begin{align}\label{eq:cov-gcsm}
        \begin{split}
            k\gcsmij
            &=\invF\big[\freq{g}_{{\gcsm},i}({\vs})\cdot\upline{\freq{g}_{{\gcsm},j}}({\vs})\big](\vtau)\\
            &={c_{ij}}\exp\big(-\frac{\vtau_{\stime}\tra{{\Var}_{ij}}\vtau_{\stime}}{2}\big)\exp\big(\imath\,({\vtau_{\stime}\tra{\vmu}_{ij}}-\vphase_{ij}\pi)\big),
        \end{split}
    \end{align}
    where $\vtau_{\timedelay}\triangleq2\pi(\vtau-\frac{{\vtime}_{ij}}{2})$ is the Euclidean distance with time delay and
    $c_{ij}={w_{ij}}{a_{ij}}$ is the normalization term incorporating the cross weight and cross amplitude and it does not depend on $\vtau$. 
    Note that $c_{ij}$ indicates the largest degree of the dependency structure because the exponential term has a max value of 1.
    
    Given an SM kernel with $Q$ components, we can obtain the corresponding dependency structures by considering the symmetric properties of SD as follows:
    \begin{align}\label{eq:SMD}
        \begin{split}
            k_{\gcsm}
            &=\sum_{i=1}^{Q}\sum_{j=1}^{Q}{c_{ij}}
                \exp\big(-\frac{\vtau_{\stime}\tra{{\Var}_{ij}}\vtau_{\stime}}{2}\big)
                \cos({\vtau_{\stime}\tra{\vmu}_{ij}}-\vphase_{ij}\pi).
        \end{split}
    \end{align}
    
    The positive semidefinite (PSD) property of the SMD kernel is equivalent to saying that its SD, ${\freq{k}_{{\gcsm}}({\vs})}$, is PSD as well \cite{Bochner2016,stein2012interpolation}. 
    Given any finite set of non-zero vectors $[{\vz}_{1}, ..., {\vz}_{N}]\tra\in\mathbb{C}^{N\times{P}}$ with complex entry, ${\vs}\in{\bbR}^{{P}}$, we have $\sum_{n=1}^{N}{\big{|}\sum_{i=1}^{Q}{\vz}_{n}\freq{g}_{{\gcsm},i}({\vs})\big{|}}^{2}\geq{0}$.
    Hence, 
    the SMD kernel must be PSD. 
    We have $\cov(f_{\sm, i}, f_{\sm, j})= k{\gcsmij}(\vtau)$ to represent the dependency structure in \eref{cov-dep}.  
    There are $Q^2$ structures     
    with $Q$ original components    
    plus $Q^2-Q$ dependency structures.
    
    \tbf{Quantification of the dependency structure:}
    To measure the intensity of the dependency structure, we normalize the dependency structure as
    \begin{align}\label{eq:learned-corr}
        \gamma_{ij}(\vtau) = \frac{k_{\gcsm}^{\itimej}(\vtau)}{\sqrt{w_{i}k_{{\sm},i}(\vtau)\cdot w_{j}k_{{\sm},j}(\vtau)}}.
    \end{align}
    
    Note that $\gamma_{ij}$ has a range with $[-1, 1]$. For $i=j$, we have $\gamma_{ij}=1$ when $w_{i}k_{{\sm},i}(\vtau)>0$ and $\gamma_{ij}=-1$ when $w_{i}k_{{\sm},i}(\vtau)<0$. 
    
    \subsection{Interpretation of dependency structure}   
    
    In \fref{sm-samp},  we show the covariances, SDs, sampling paths, and posterior distributions in terms of amplitude, peak, and trend between the SM (dashed red) and SMD (dashed blue) kernel. The differences between SM and SMD are clear. 
    Without TP delays, subplots (b) and (g) of \fref{sm-samp} show that the dependency structure can reinforce the magnitudes of both SD and covariance in SM (shown in subplots (a) and (f)) but does not change the decaying behavior of covariance a lot. The dependency structure in the frequency domain (subplot (g)) is the intersection (modeled as a Gaussian, see \eref{cross_gcsm_spec}) between two SM components.
    When having TP delays, the dependency structure can reinforce or weaken the covariances and SDs of the original kernel. In subplots (c) and (e), the covariance range of SMD is much extended and larger than SM due to the time delay.
    Specifically, the dependency structure can largely change the magnitudes of both SD (shown in subplots (h), (i), and (j)) and covariance (shown in subplots (c), (d), and (e)), shapes of SD, and decaying behaviors of the covariance, and further reduce the predictive uncertainties (show in subplots (m), (n), and (o)).
 
    Given six observations (marked with black crosses) and conditions on them, the learned posterior distribution and sampling path are shown in subplots (k), (l), (m), (n), and (o). Interestingly, due to the dependency structure, the predictive confidence interval (CI) of the SMD is significantly tighter (in blue shadow) than that of the SM (in red shadow).

    \begin{figure*}[h!]
        \centering
        \scriptsize
        \renewcommand{\tabcolsep}{1.5mm}
        \def\figart#1#2{\includegraphics[width=#1\columnwidth]{fig-#2.pdf}} 
        \begin{tabular}{p{1.5mm}*{3}{c}} 
            & \figart{0.280}{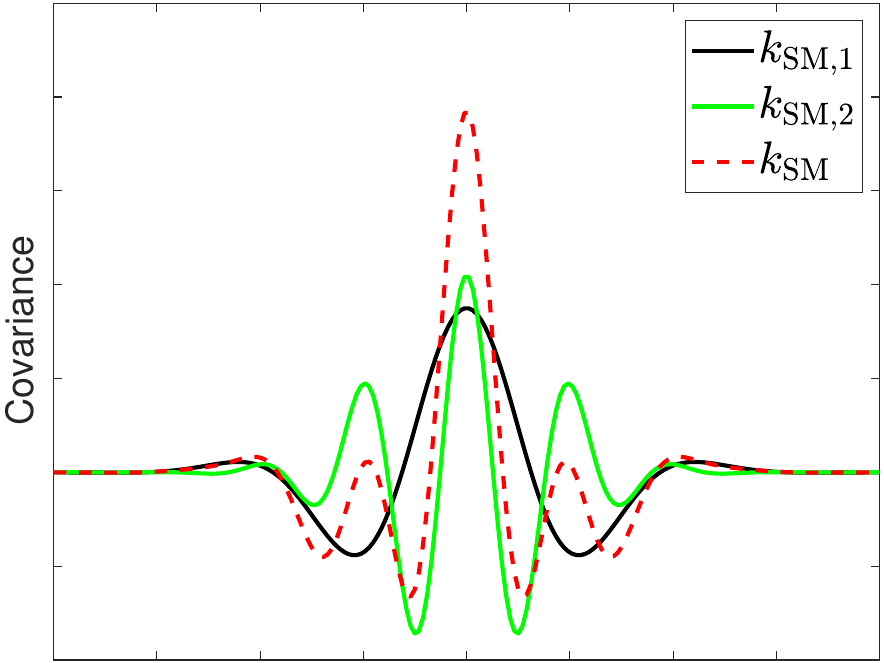} 
            & \figart{0.280}{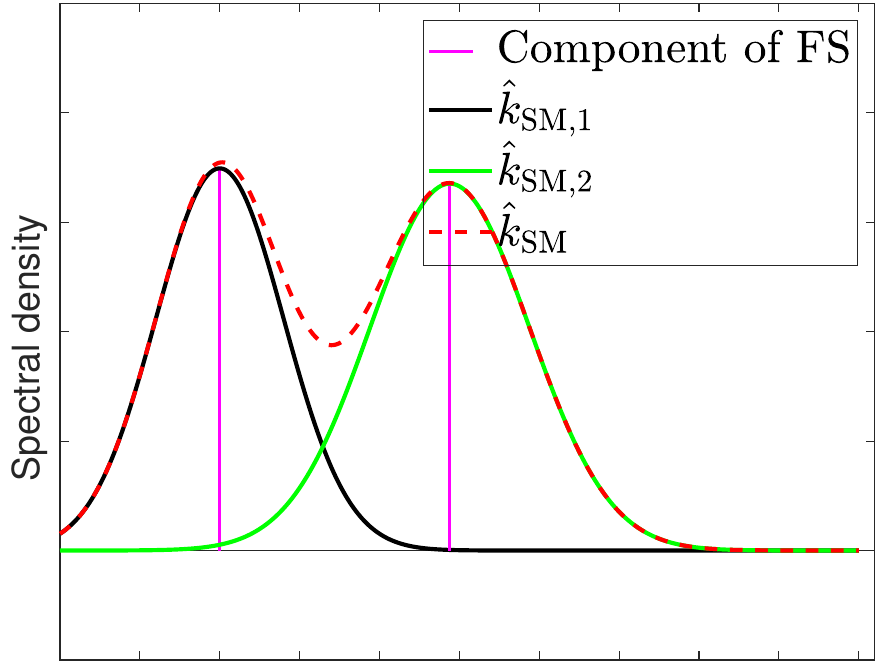} 
            & \figart{0.280}{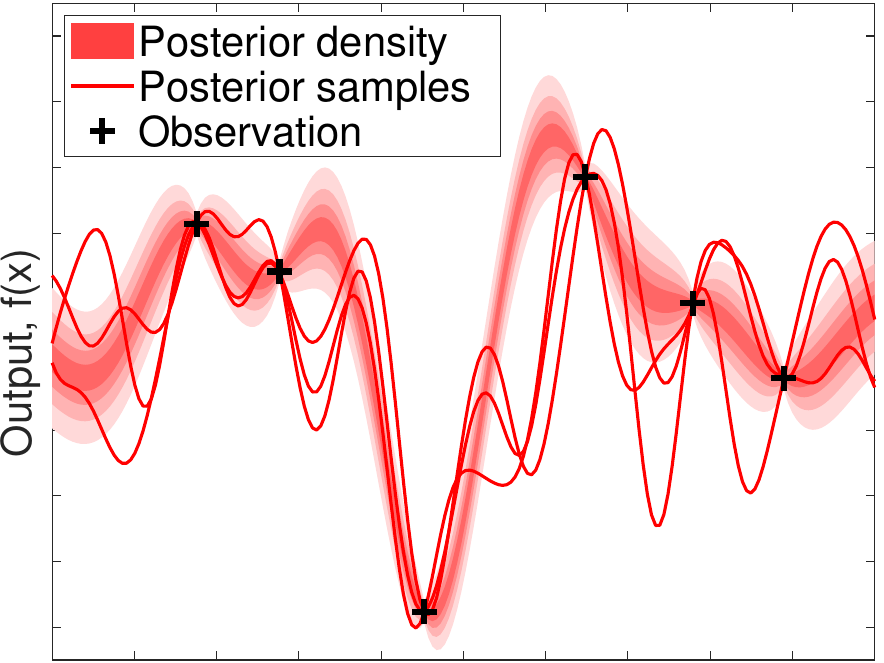} \\
            & (a) $k_{\sm}$ & (f) $\freq{k}_{\sm}$ & (k) $f_{\sm}$    \\
            & \figart{0.280}{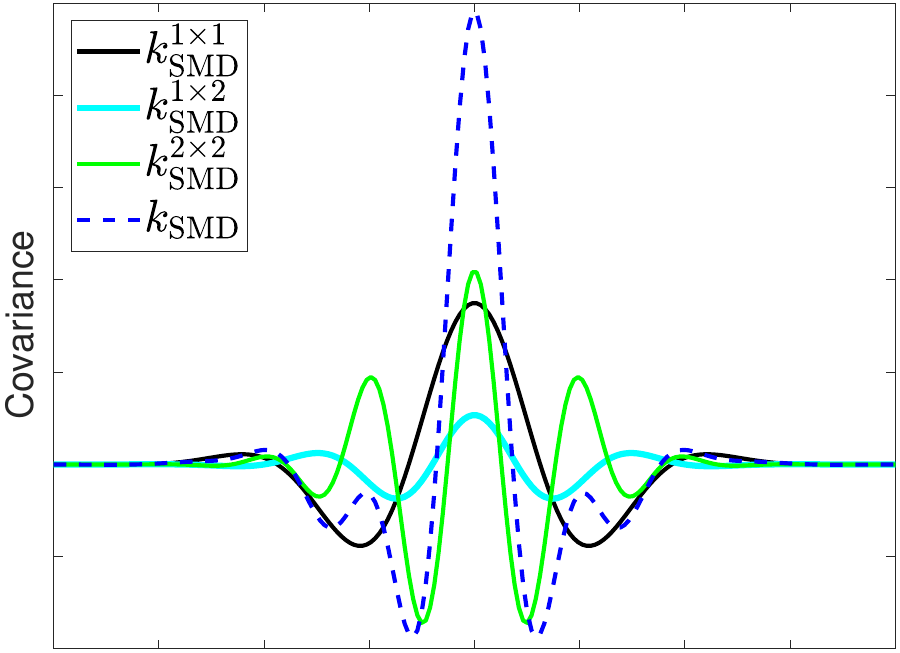} 
            & \figart{0.280}{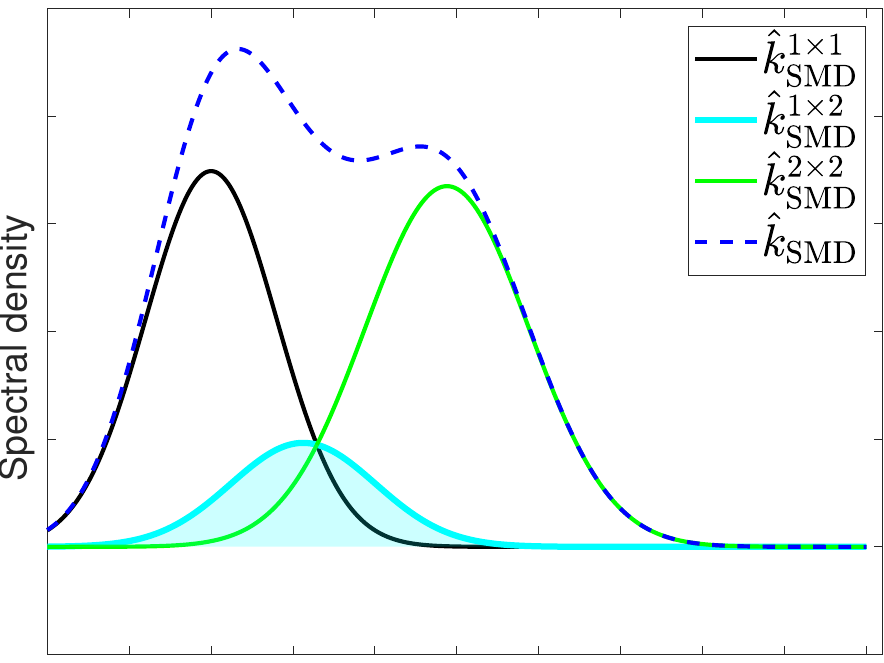} 
            & \figart{0.280}{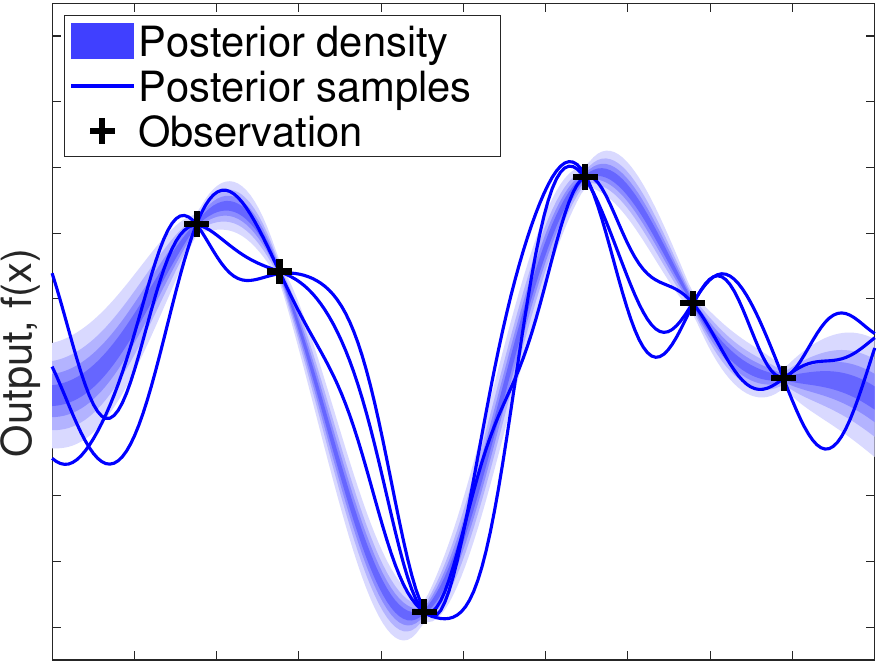} \\
            & (b) ${k}_{\gcsm}(\stime=0, \sphase=0)$ & (g) $\freq{k}_{{\gcsm}}(\stime=0, \sphase=0)$ & (l) $f_{{\gcsm}}(\stime=0, \sphase=0)$ \\
            & \figart{0.280}{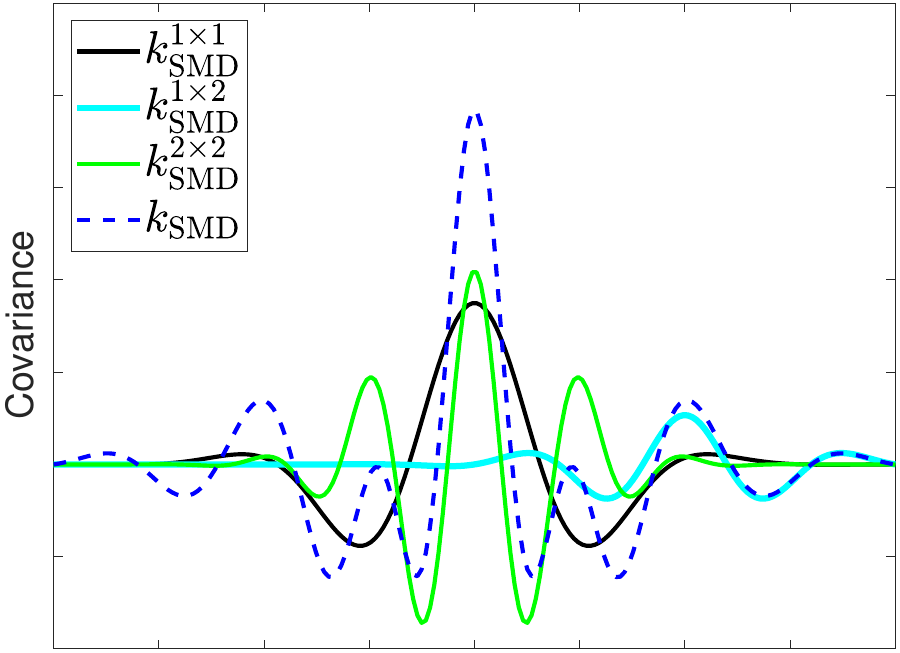} 
            & \figart{0.280}{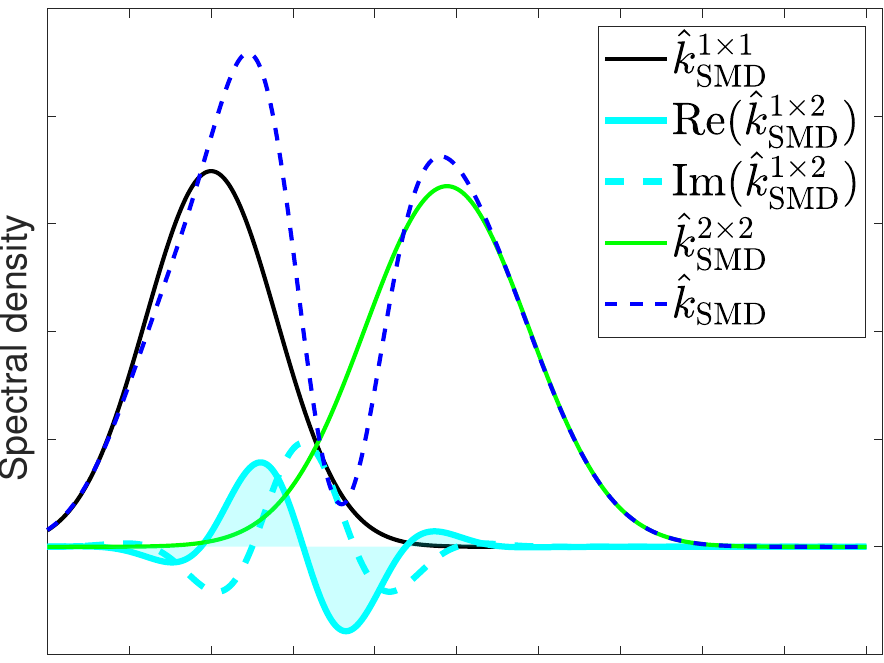} 
            & \figart{0.280}{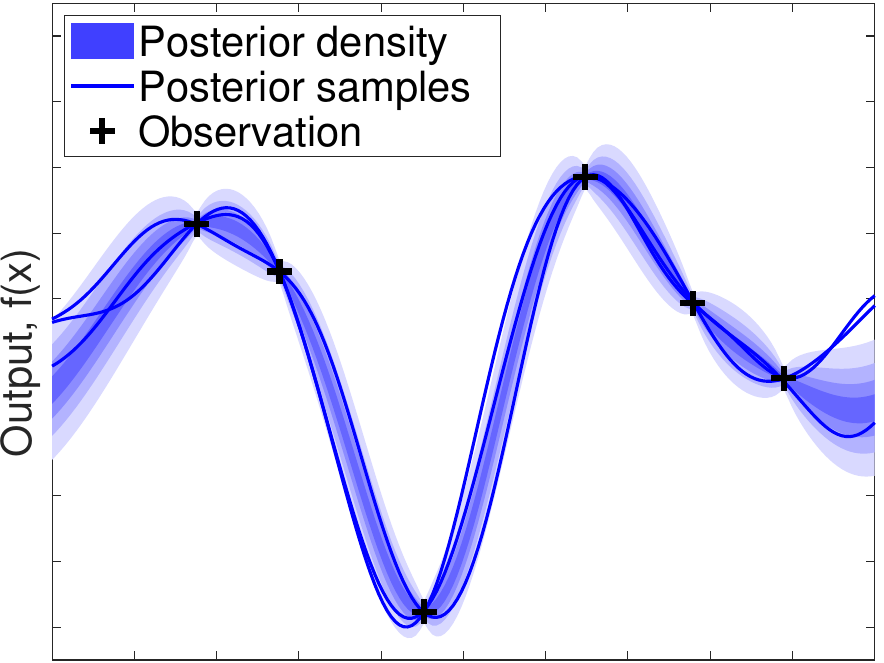} \\
            & (c) ${k}_{{\gcsm}}(\stime\neq{0}, \sphase={0})$ & (h) $\freq{k}_{{\gcsm}}(\stime\neq{0}, \sphase={0})$ & (m) $f_{{\gcsm}}(\stime\neq{0}, \sphase={0})$ \\
            & \figart{0.280}{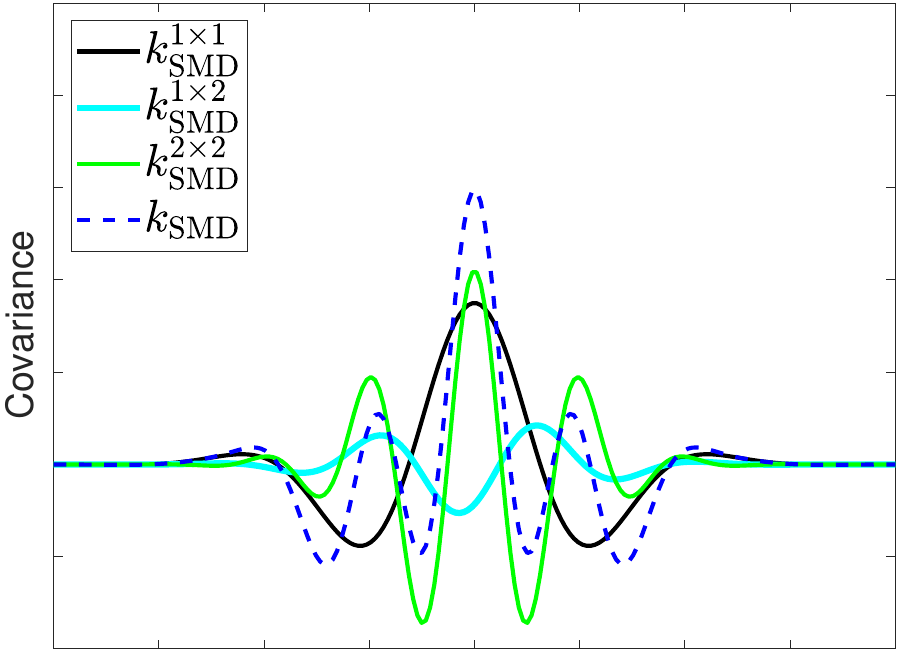} 
            & \figart{0.280}{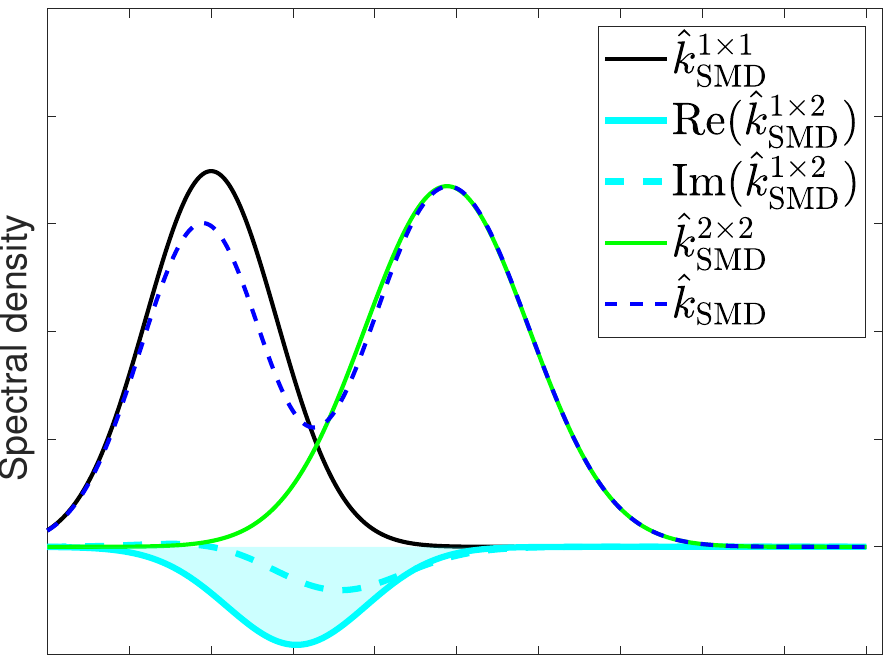} 
            & \figart{0.280}{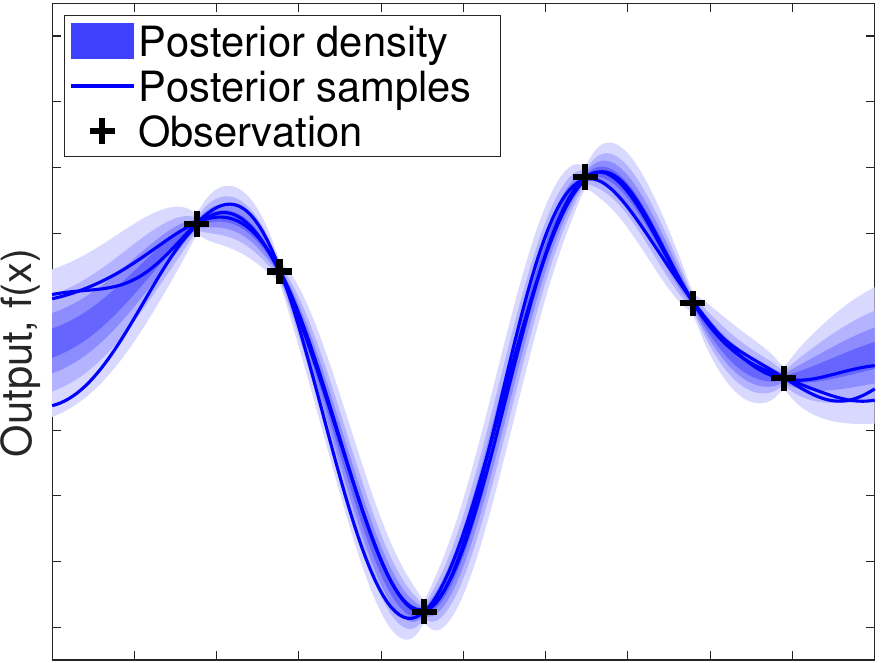} \\
            & (d) ${k}_{{\gcsm}}(\stime=0, \sphase\neq{0})$ & (i) $\freq{k}_{{\gcsm}}(\stime=0, \sphase\neq{0})$ & (n) $f_{{\gcsm}}(\stime=0, \sphase\neq{0})$ \\
            & \figart{0.280}{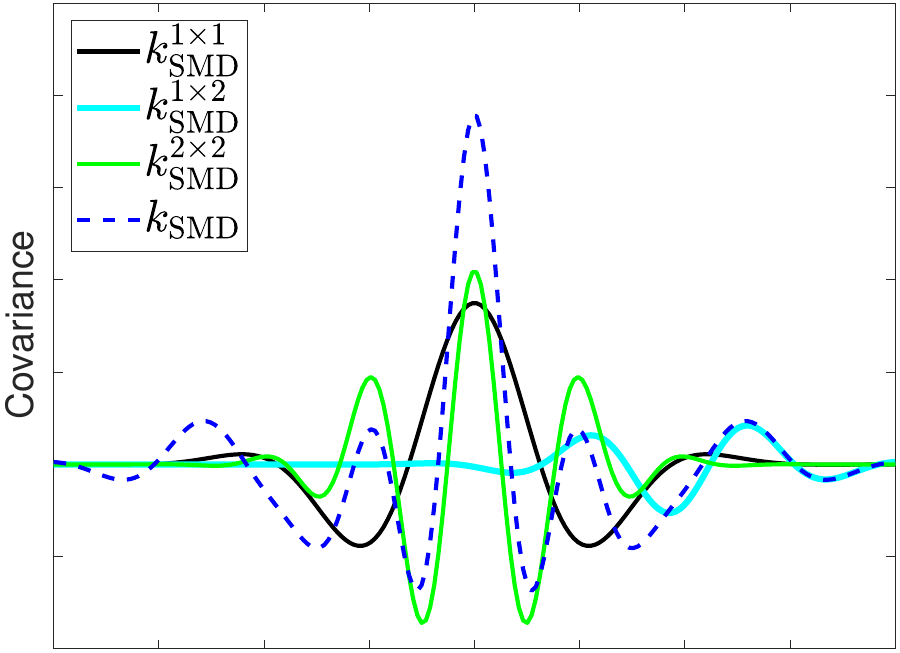} 
            & \figart{0.280}{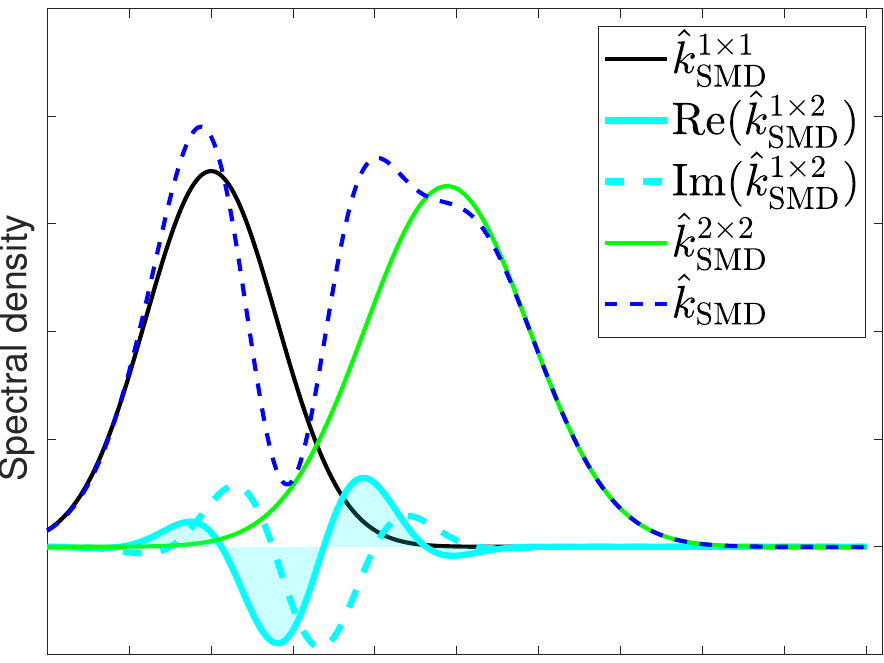} 
            & \figart{0.280}{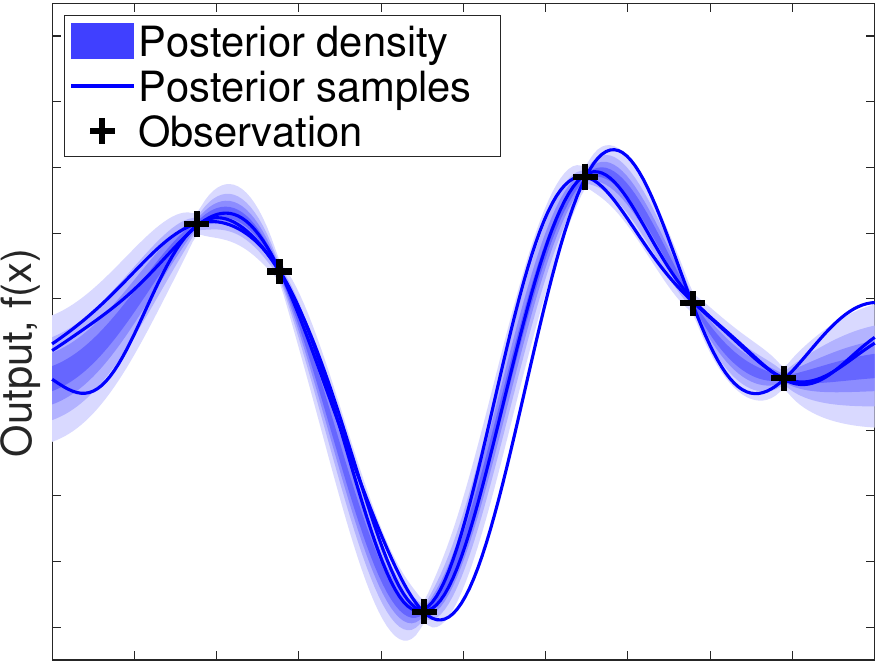} \\
            & (e) ${k}_{{\gcsm}}(\stime\neq{0}, \sphase\neq{0})$ & (j) $\freq{k}_{{\gcsm}}(\stime\neq{0}, \sphase\neq{0})$ & (o) $f_{{\gcsm}}(\stime\neq{0}, \sphase\neq{0})$ \\
        \end{tabular}
        \caption{Covariances (the first column), SDs (the second column), sampling path (the third column), and posterior distributions based on GPs with the SM ($Q=2$) and SMD ($Q=2$) kernels conditioned on six observations.
            The samples of all GP models were obtained using 200 equally spaced points. 
        }
        \label{fig:sm-samp}
    \end{figure*}   
    
    \subsection{Comparisons between the SMD and related kernels}\label{sec:comp}
    
    \begin{figure}[h!]
        \def\figexp#1{\includegraphics[width=#1\columnwidth, clip, trim=0.1cm 1.5cm 0.0cm 0.0cm]{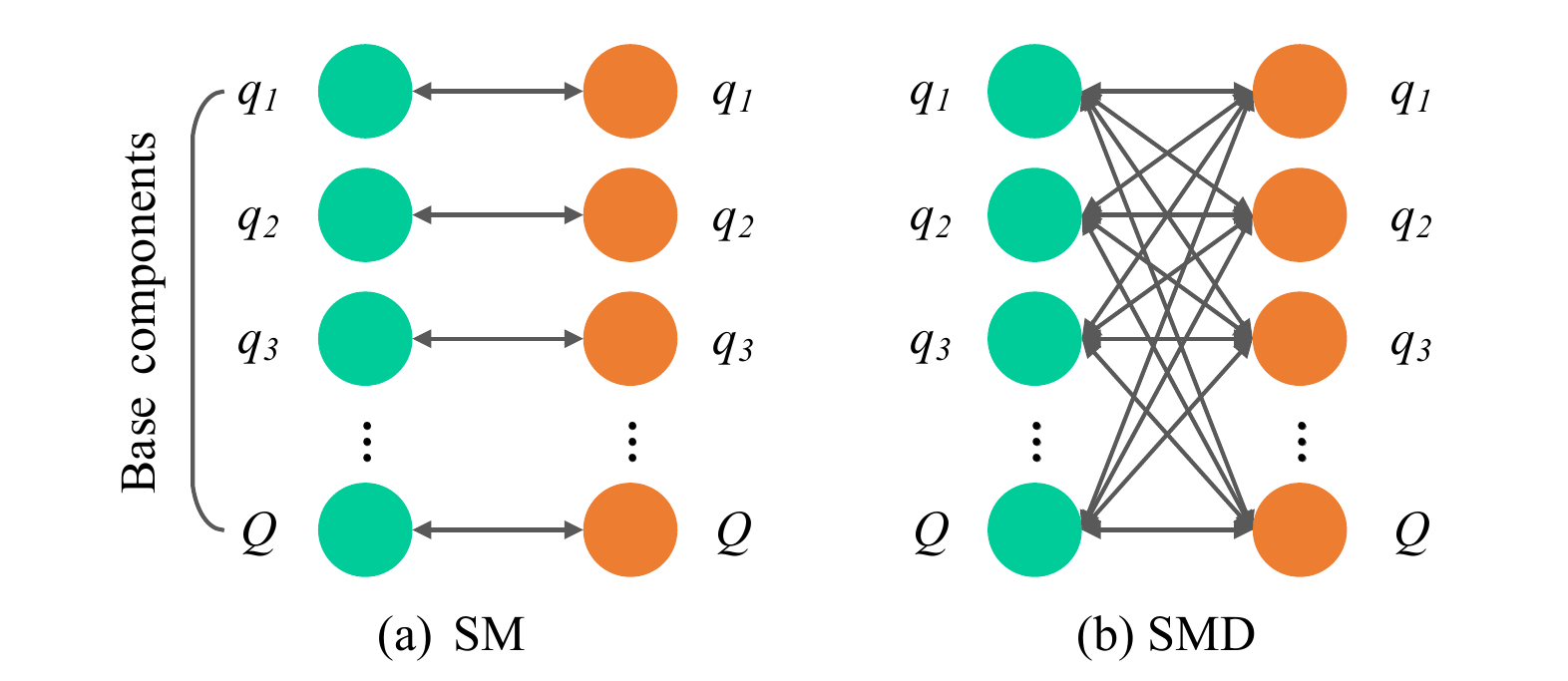}}
        \begin{center}
            \figexp{0.6}{}
            \caption{SM kernel (left) vs SMD kernel (right) with $Q$ original components, where $q_{i=\{1,..., Q\}}$ denotes the $i$-th SM component. The SM models only the autocovariance between the component itself. The SMD models both auto- and cross-covariance between components.}
            \label{fig:frame}
        \end{center}
    \end{figure}
    
    In \fref{frame}, we visualize the covariance differences between the SM and SMD (shown in \eref{cov-dep}), where each link (in black solid) represents a covariance structure of the kernel. Circle $q_{i}$ corresponds to a $f_{\sm, i}$. The cross connection denotes a $\cov(f_{\sm,i}, f_{\sm, j})$ of $f_{\sm, i}$ and $f_{\sm, j}$. 
    The SM considers only the autocovariance $\cov(f_{\sm,i}, f_{\sm, i})$ of its components and ignores their dependency structures. 
    \tref{hyp} summarizes the differences between the SMD and SM kernels in terms of their hyperparameters for a $P$-dimensional input setting. 
    For NSM,
    each hyperparameter of the original SM is parameterized as a GP with a squared exponential (SE) kernel, for instance, the weight $w_{i}$ becomes $w_{i,\vx}\sim\gp
    (0, k_{\se}(\vx, \vx'))$ in NSM. Thus NSM needs three times more hyperparameters than SM because $w_{i,\vx}$ usually has three hyperparameters.
    Without TP delays, the hyperparameter space of the SMD is equal to that of the SM.
    The price paid for incorporating TP delays in the SMD is that the gradient computation is more involved because of additional TP hyperparameters.
    
    \begin{table}[h!]
        \scriptsize
        \caption{Comparisons between the SMD and other SM kernels in terms of hyperparameters and the number of hyperparameters.  For an initial large $Q$ in the SM and SMD, the number of components retained after compression, $Q_{\rest}$, is much smaller than $Q$.
        }
        \begin{center}
            \begin{tabular}{ccc}
                \toprule
                {Kernel} & {hyperparameters} & {Number of hyperparameters} \\
                \midrule
                SM & $\{w_{i},\,\vmu_{i},\,{\Var}_{i}\,\}_{i=1}^{Q}$ & $(2P+1){Q}$ \\
                NSM & $\{w_{i,\vx},\,\vmu_{i,\vx},\,{\Var}_{i,\vx}\,\}_{i=1}^{Q}$ & $3\times(2P+1){Q}$ \\
                SMD$_{\vphase=0, \vtime=0}$ & $\{w_{i},\,\vmu_{i},\,{\Var}_{i}\}_{i=1}^{Q}$ & $(2P+1){Q}$\\ 
                \tbf{SMD}  & $\{w_{i},\,\vmu_{i},\,{\Var}_{i},\,{\vtime}_{i},\,{\vphase}_{i}\}_{i=1}^{Q}$ & ${(4P+1){Q}}$ \\
                \tbf{Compressed SMD} & $\{w_{i},\,\vmu_{i},\,{\Var}_{i},\,{\vtime}_{i},\,{\vphase}_{i}\}_{i=1}^{Q}$ & ${(4P+1){Q_{\rest}}}$ \\
                \tbf{SMD with SA} & $\{w_{i},\,\vmu_{i},\,{\Var}_{i},\,{\vtime}_{i},\,{\vphase}_{i}\}_{i=1}^{Q}$ & ${(1+2P(2-\alpha_\sr)){Q_{\rest}}}$ \\
                \bottomrule
            \end{tabular}
        \end{center}
        \label{tab:hyp}
    \end{table}
    
    \section{Structure adaptation for the spectral mixture with dependency structure}\label{sec:hyp}
    
    
    \begin{algorithm}
        \small
        \SetAlgoLined
        \SetKwInOut{Input}{Input}
        \SetKwInOut{Output}{Output}
        
        \SetAlgoLined
        \Input{Initial number of components, $Q_{\init}$,             
            number of training attempts, $M_\init$,
            number of pruned components, $Q_{\prune}=0$.
        }
        \Output{Fine-tuned sparse GP with SMD kernel.
        }
        \BlankLine    

        Pretrain $\gp
        (0, k_{\gcsm}(Q_{\init}))$ (initialized by BHI, \aref{hypinit}) $M_\init$ times\;
        Choose $\tilde{\Theta}_{\best}$ with the lowest $\loss$ from pretrainings\;   
        \For{all $k\gcsmij$ ($i=j$) }{
            Obtain the $i$-th weight $w_{i}$ from $\tilde{\Theta}_{\best}$\;
            \If{$w_{i}<1$}{
                Remove the $i$-th component in SMD\;
                Remove $\{w_{i}, \vmu_{i}, \Sigma_{i}, \vtime_{i}, \vphase_{i}\}$ in $\tilde{\Theta}_{\best}$\;
                $Q_{\prune}=Q_{\prune}+1$\;}}
        Remove the low intensity dependency structures with $c_{ij}<1$\;
        Fine training the GP with the sparse SMD kernel\;
        \caption{Structure adaption for SMD}
        \label{algo:SMD}
    \end{algorithm}
    
    The SM kernel has been known for its large number of hyperparameters (with size $3Q$) \cite{Wilson2013}. This complicates the inference, learning, and interpretability of GPs with the SM kernel. Critically, several de facto inference and learning issues impede the use of the SM kernel, such as hyperparameter initialization and choosing the number of kernel components.
    The SMD also suffers from these issues. 
    The dependency structures in SMD are dense and therefore need to be sparsified. 
    We propose a structure adaptation (SA) algorithm (see \aref{SMD}) for the SMD to handle the above issues and to achieve efficient inference and interpretable structure discovery.
    In \aref{SMD}, the symbol $\tilde{\Theta}_{\best}$ denotes inferred hyperparameters of better training. Steps 
    1-2 perform $M_\init$ trainings
    to obtain a better hyperparameter position with a smaller loss.
    Steps 3-10 prune the unimportant components by comparing their weights. Steps 11-12 sparsify the dependency structures by quantifying their intensity, removing the weak ones, 
    and fine-train the GP with sparse dependency structures. 
    In \aref{SMD}, we use a standard maximum-likelihood approach for the optimization (estimation) of hyperparameters. Such optimization is performed in pretrain stage (step 1) and fine training stage (step 12) of \aref{SMD}. 
    We have two levels of sparsity for the SMD kernel: the first level of sparsity is obtained from the compression of original components and the second level of sparsity is handled by the reduction of weak dependency structures. 
    Specifically, we introduce the details of the proposed SA algorithm in the following subsections. 
    
    
    \subsection{Bootstrap-based hyperparameter initialization (BHI)}\label{sec:hyp_init}
    
    The learning of the SMD kernels relies on a good starting point when performing optimization in high-dimensional hyperparameter space. A better initialization can help us more easily discover the underlying structure.
    Sniffing the structure of the empirical SDs can alleviate the difficulty of hyperparameter initialization due to the connection (indicated by Bochner's Theorem) between the SD and kernel \cite{Wilson2013,Herlands2016}.
    However, the empirical SDs are biased estimates of the true underlying spectral structures, which contain noise and fake peaks denoting spurious patterns. 
    To filter out the noise and fake peaks, we employ the bootstrap techniques \cite{zoubir1998bootstrap} to improve the estimation accuracy of the empirical SDs.
    We draw a large number of spectral samples $S^{*}$
    using bootstrap with replacement from the empirical SDs. 
    We then consider a Gaussian mixture model (GMM) fitting to the 
    bootstrap samples $S^{*}$  to obtain the $Q$ Gaussians, ${p}(\tilde{\Theta} |\vs)=\sum _{i=1}^{Q}{\tilde {w_{i}}}\N(\vs; \tilde{\vmu}_{i},{\tilde {\Sigma} _{i}})$.
    Finally, we propose a BHI algorithm (shown in \aref{hypinit}) for the SMD. 
    The bootstrap sampling times $B=100$ are generally sufficient for robust statistics estimation.
    The estimates of the hyperparameters obtained in bootstrap are used for initialization.  The \aref{hypinit} is performed before optimization in pretrain stage (step 1) of \aref{SMD}. 
    An illustration of \aref{hypinit} is shown in \fref{bootstrap-init}. 

    \begin{algorithm}
        \small
        \SetAlgoLined
        \SetKwInOut{Input}{Input}
        \SetKwInOut{Output}{Output}
        
        \SetAlgoLined
        \Input{$Q_{\init}$, $B=100$.}
        \Output{Hyperparameter initialization $\tilde{\Theta}_{\init}$.}
        \BlankLine    
        Compute the empirical SD $S$ using the Blackman window and FT\;
        Resample bootstrap spectral samples $S^{*}\subset{S}$ 
        from $S$
        \;
        Fit a GMM with $Q_{\init}$ components to $S^{*}$ and obtain a bootstrap estimate $p(\tilde{\Theta}^{*} |\vs)$\;
        Sort $Q_{\init}$ components with mean position $\tilde{\vmu}_{i}^{*}$ in $\tilde{\Theta}^{*}$\;
        Repeat steps 2-4 $B$ times to obtain $B$ estimates $p(\tilde{\Theta}^{*}_{1} |\vs), p(\tilde{\Theta}^{*}_{2} |\vs),..., p(\tilde{\Theta}^{*}_{B} |\vs)$\;
        The final bootstrap estimates of the hyperparameters are computed as 
        $\tilde{\Theta}_{\init}=\frac{1}{B}\sum _{i=1}^{B} \tilde{\Theta}^{*}_{i}$.
        \caption{Bootstrap-based hyperparameter initialization}
        \label{algo:hypinit}
    \end{algorithm}
    
    \begin{figure}[h!]
        \centering
        \renewcommand{\tabcolsep}{-0.0mm}
        \def\figart#1#2{\includegraphics[width=#1\columnwidth]{fig-GCSM-bootstrap-#2.pdf}}
        \begin{tabular}{p{0.5mm}*{1}{c}}
            & \figart{0.5}{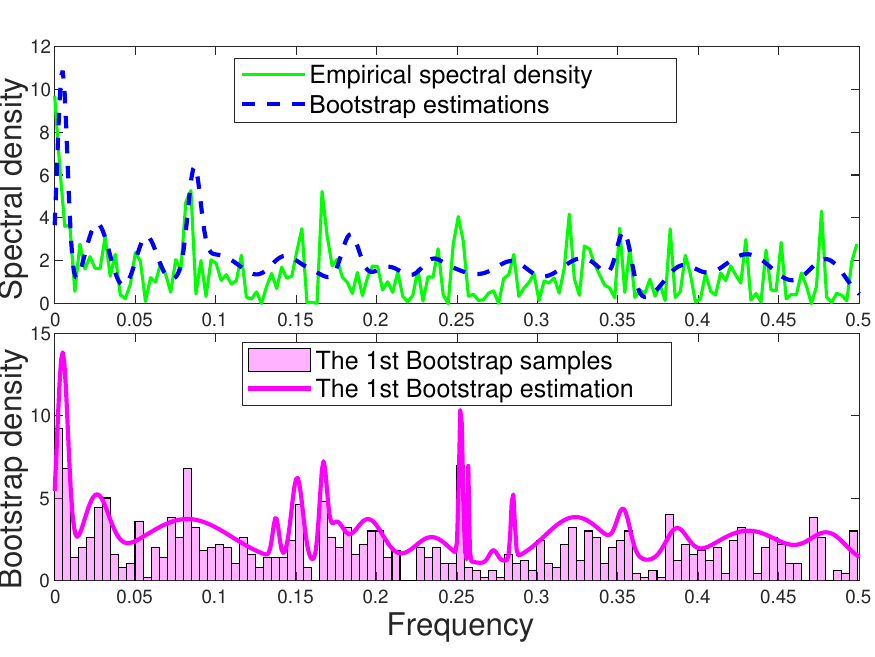}
        \end{tabular}
        \caption{The BHI algorithm on the monthly river flow dataset. The first subplot: the empirical SD (in green line) and bootstrap estimation (in dashed blue line). The second subplot: the 1st bootstrap samples (in magenta bar) and the corresponding estimation (in magenta line). Many small peaks in empirical SD are filtered in the bootstrap estimation. }
        \label{fig:bootstrap-init}
    \end{figure}
    
    
    \subsection{Compressed spectral mixture with dependency structures}\label{sec:prune}
    How to set the number of components is another challenging issue for SM and SMD kernels. We must specify the $Q$ in advance and fix it during optimization. However, inaccurate $Q$ cannot reflect the true number of underlying patterns contained in data, which could lead to overfitting for large $Q$ or underfitting for small $Q$. This makes all spectral kernels flawed for real-world applications. 
    
    To adaptively select the number of components, we first prune the minor components by quantifying their weights. 
    As shown in \fref{prune-river1}, we demonstrate the learned importance of components in the SM kernel by using the monthly river flow dataset.
    Specifically, the weights of components ($i=\{1, 2, 3, 6, 8, 9\}$) in the left subplot are smaller than 1, which means that the corresponding amplitudes in the frequency domain are pretty small. Observing this fact, we simply think a component with a weight smaller than 1 is less important and takes a tiny portion of the signal energy.
    
    \begin{figure}[h!]
        \centering
        \scriptsize
        \renewcommand{\tabcolsep}{-0.0mm}    \def\figart#1#2{\includegraphics[width=#1\columnwidth]{fig-GCSM-#2.pdf}}    
        \begin{tabular}{p{0.5mm}*{2}{c}}
            & \figart{0.40}{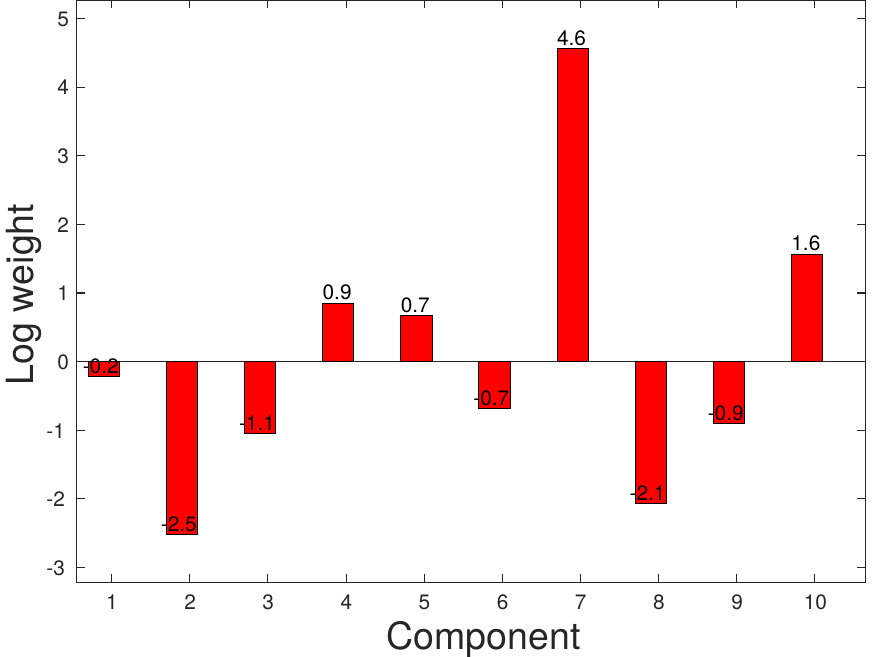}
            & \figart{0.44}{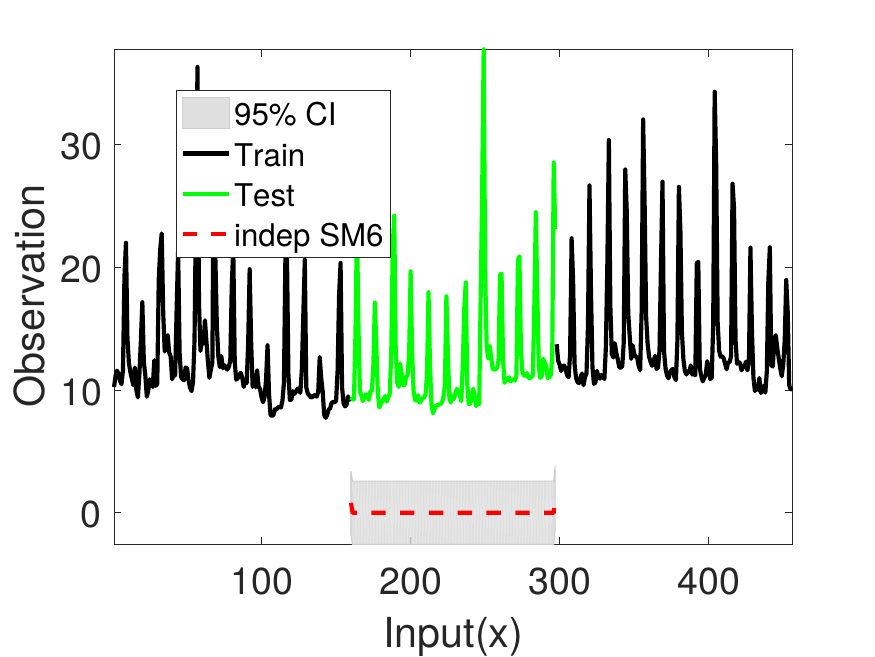} \\
        \end{tabular}
        \caption{The learned $w_{i}$ (left) of SM kernel and the interpolation result (right) of the $6$-th SM component with a smaller weight. Here, the predictive mean of the $6$-th SM component is almost zero and contributes a little to the final predictive distribution and thus can be pruned.}
        \label{fig:prune-river1}
    \end{figure}
    
    Removing such components with small weights does not affect a GP model's learning and generalization ability.
    Consequently, we propose a pruning strategy to compress the SMD. 
    The pruning strategy is described by steps 2-9 of \aref{SMD}. 
    This strategy can reduce the number of hyperparameters in SMD to $5Q_{\rest}$, where $Q_{\rest}$ is the rest components after compression. We define a CR for the SMD, $\alpha_\compressr = 1 - \frac{Q_{\rest}}{Q_{\init}}\times{100\%}$, to assess how much the SMD is compressed. 
    \begin{remark}
        $\alpha_\compressr$ is an indicator of the pruning degree of the SMD using the SA algorithm. In the extreme cases, $\alpha_\compressr$ is \%100 if all components are pruned with $Q_{\rest}=0$ and $100\%$ if all components are kept with $Q_{\rest}=Q_{\init}$. 
    \end{remark}
    
    
    \subsection{Sparse dependency structure and its behavior}\label{sec:sparse-dep}
    In this section, we investigate the sparsity and behavior of the dependency structures in SMD. 
    Observing from \eref{cov-dep} and \eref{SMD}, there are $Q^{2}-Q$ dependency structures.
    In fact, for two components located far away from each other, the intersection between their SDs is close to zero, which means that their dependency is weak.
    \eref{cross_gcsm_spec} indicates that the closer the $\vmu_{i}$, $\Sigma_{i}$ and $w_{i}$ between components are, the greater the dependency is, and vice versa. Hence, we can confidently remove the low-intensity dependency structures.
    
    Specifically, 
    we introduce a binary mask ${\beta_{ij}}$ determined by $w_{ij}$ and $a_{ij}$ to indicate whether remove the dependency structure between components $i$ and $j$. 
    We have the compressed SMD with sparse dependency structures as 
    \begin{align}\label{eq:SMD-SA}
        \begin{split}
            {k_{\gcsm}^{\sa}(\vtau)}=\sum_{i=1}^{Q_{\rest}}\sum_{j=1}^{Q_{\rest}}&\beta_{ij}c_{ij}\exp\big(-\frac{1}{2}{\vtau_{\stime}\tra{{\Var}_{ij}}\vtau_{\stime}}\big)\cos({\vtau_{\stime}\tra{\vmu}_{ij}}-\vphase_{ij}\pi),
        \end{split}
    \end{align}
    where  
    ${\beta_{ij}}=0$ if $c_{ij}<1$ and $i\neq{j}$; otherwise, ${\beta_{ij}}=1$.

    \fref{sm-samp} shows the neat contribution of the dependency structure to the final covariance even with zero TP delays. When ${\stime}\neq0$ or ${\sphase}\neq0$, the covariance (in cyan) of the dependency structure is shifted and centered at a different position (shown in subplots (c), (d), and (e) of \fref{sm-samp}). We define an SR of the dependency structure for the SMD, $\alpha_\sr =\Big(1-\frac{\sum_{i=1}^{Q_{\rest}}\sum_{j=1}^{Q_{\rest}}{\beta_{ij}}}{Q^{2}_{\rest}-Q_{\rest}}\Big)\times 100\%,$
    to evaluate how sparse the dependency structure is. 
    \begin{remark}
        The $\alpha_\sr$ is ensured to be in the range of $[0, 1]$.
        Note that $\alpha_\sr$ is $1$ if there is no significant dependency structure or $0$ if all dependency structures are large.
    \end{remark}
    
    
    \section{Experiments}\label{sec:exp}
    In this section, we comprehensively investigate the performance of the SMD and compare it with that of some state-of-the-art kernels on both synthetic and real-world datasets. For all experiments, the popular kernels implemented in the GPML toolbox \cite{Rasmussen2006} are used as baselines, such as the linear (LIN), SE, polynomial (Poly), periodic (PER), rational quadratic (RQ), Mat\'ern (MA), Gabor, fractional Brownian motion covariance (FBM), underdamped linear Langevin process covariance (ULL), neural network (NN) and SM kernels. 
    The same number of components $Q$
    is used for the SM, NSM, and SMD kernels. 
    In all plots, the training data, testing data, SM prediction, SMD prediction, 
    and CI are shown in black, green, red, blue, and gray shadow, respectively. 
    
    
    \subsection{Model assessment}
    
    We consider multiple metrics to assess the performances and characteristics of GP models, such as 
    \begin{itemize}
        \item the mean squared error (MSE) defined as ${\mathrm {MSE}\triangleq{\frac{1}{n}\sum _{i=1}^{n}\big(y_{i}^{*}-\tilde{y}_{i}^{*}\big)^{2}}}$ to measure the generalization performance of GP;   
        \item the 95\% CI (instead of, e.g., error bar) to visualize the uncertainty
        of prediction;
        \item 
        the posterior correlation $\rho_{ij}$ 
        to quantify the latent dependency between SM components;
        \item the $\gamma_{ij}$ (see \eref{learned-corr}) 
        to indicate  
        the intensity of dependency structure learned by the SMD;
        \item the CR ($\alpha_{\compressr}$) and SR ($\alpha_{\sr}$) of the SMD.
    \end{itemize}
    
    
    \subsection{Learning a synthetic signal with dependency structure}\label{sec:syn}
    
    
    \begin{figure}[h!]
        \centering
        \renewcommand{\tabcolsep}{0.5mm}
        \def\figart#1#2{\includegraphics[width=#1\columnwidth, clip, trim=0.0cm 0.0cm 0.5cm 0.0cm]{fig_GCSM_IntervalSM-#2.pdf}}
        \begin{tabular}{p{0.1mm}*{2}{c}}
            & \figart{0.4}{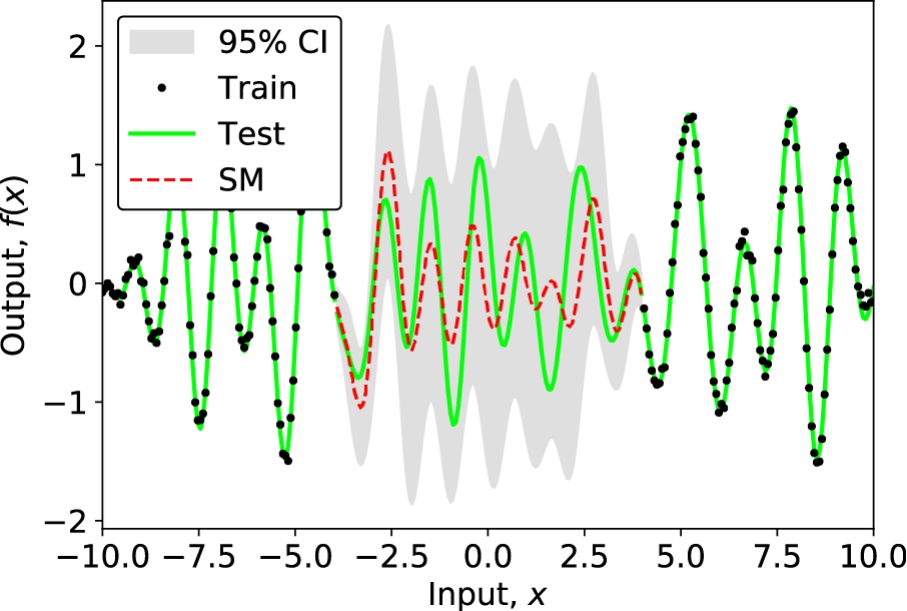}
            & \figart{0.4}{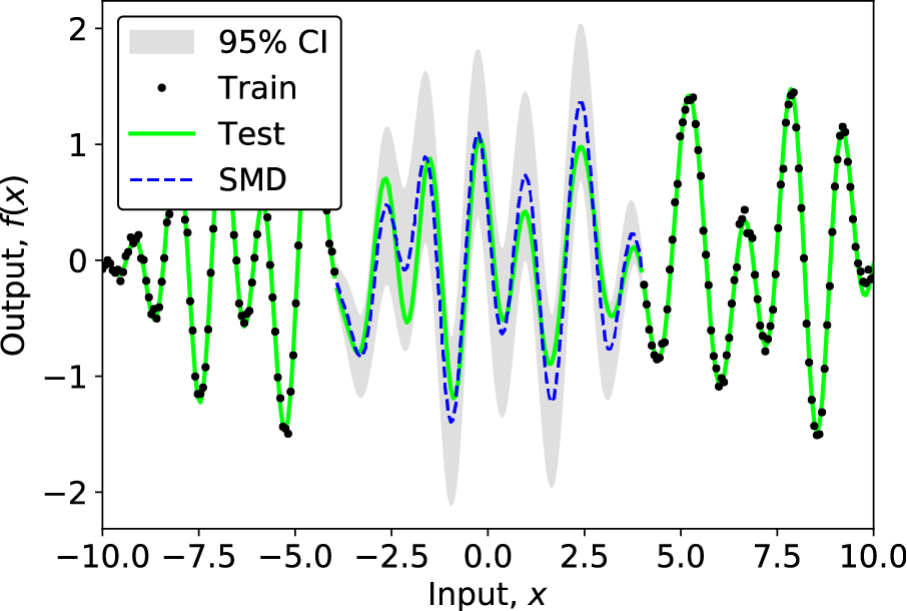} \\
        \end{tabular}
        \caption{Performance of the SM (left) and SMD (right) on a synthetic signal. 
        }
        \label{fig:artificial-data}
    \end{figure}

    We illustrate the capability of the SMD to capture TP delayed dependency structure in a synthetic signal. The signal is sampled from the following GP with a hybrid kernel structure: $f(x)\sim\gp(0, k_{\gcsm}(\theta=\{0.1, 0.3\}, \phi=\{0.1, 0.3\})+k_{\sm})$. 
    The signal contains a dependency structure due to the employment of $k_{\gcsm}$. 
    The SM and SMD kernels of the signal $f(x)$ have $Q=2$ components. 
    We generate a time series of length 300 in the interval [-10, 10] and add some noise to it (see \fref{artificial-data}). In this experiment, we remove the middle $40\%$ of the signal and consider it as missing testing data (in green). The rest of the signal forms the training data (in black).
    Both SMD and SM are configured with 
    $Q=5$ and with the same initial values of the hyperparameters ${w_i},\,\mu_i,\,{\sigma^{2}_i}$. Other hyperparameters of SMD, ${\theta_i}$ and ${\phi_i}$, are initialized to be zeros.
    
    In \fref{artificial-data}, the performance difference between SMD and SM is clear in terms of the mean and uncertainty of the prediction. The SMD is capable of learning the hybrid covariance with dependency structure well. For the SM (dashed red), it is more difficult to recognize such a dependency structure and to interpolate the missing block. Here, the SMDs without TP delays, with only time delay, and with only phase delay cannot interpolate the missing block well. Obviously, the SMD yields better prediction and smoother CIs than the SM and therefore achieves the lowest MSE (see \tref{MSE}).
 
    \subsection{Long range interpolation of monthly river flow monitoring}\label{sec:riverflow}
    Interpolation is a well-known task for GP learning. 
    In this experiment, we validate the long-range interpolation ability of GP with the SMD kernel. We consider the monthly river flow dataset because it reveals time and phase patterns with variability.
    The moon and sun are primarily responsible for the rising and falling of tidal river flows, and such effects are delayed and augmented by gravity and resonances. The mean monthly river flow in the Madison River near West Yellowstone is the average flow from 1923 to 1960 \cite{Hipel1994}. Empirical analysis \cite{Hipel1994} shows various characteristics of this flow data experiment: short term monthly variations, medium term seasonal patterns, irregular periodic long term trends caused by the relative positions of the moon and sun, and some white noise. 
    
    As such, the monthly river flow contains complicated patterns (see \fref{riverflow}) that may be caused by physical interferences.
    The appearance time of the flow peak is not periodical, and its amplitude is always irregular. There are 456 records in the dataset. Here, $30\%$ of the data, namely, the long range middle part, is removed for testing, while the rest of the data are used for training. 

    In \tref{MSE}
    and \fref{riverflow}, the results indicate that both the SMD and SM can interpolate the missing month river flow well. 
    However, the SMD achieves better performance and confidence. The SMD is generally more effective in modeling complex patterns hidden in these data. Furthermore, using the same initial $Q$ and the SA algorithm, \tref{CR} shows that SMD achieves a better CR of 38.9\% than SM. The SR of 89.3\% indicates that most of low intensity dependency structures in the SMD are removed.

    \begin{figure}[h!]
        \centering
        \renewcommand{\tabcolsep}{0.2mm}
        \def\figart#1#2{\includegraphics[width=#1\columnwidth, clip, trim=0.3cm 0.1cm 1.0cm 0.3cm]{fig-GCSM#2.pdf}}
        \begin{tabular}{p{0.5mm}*{2}{c}}
            & \figart{0.45}{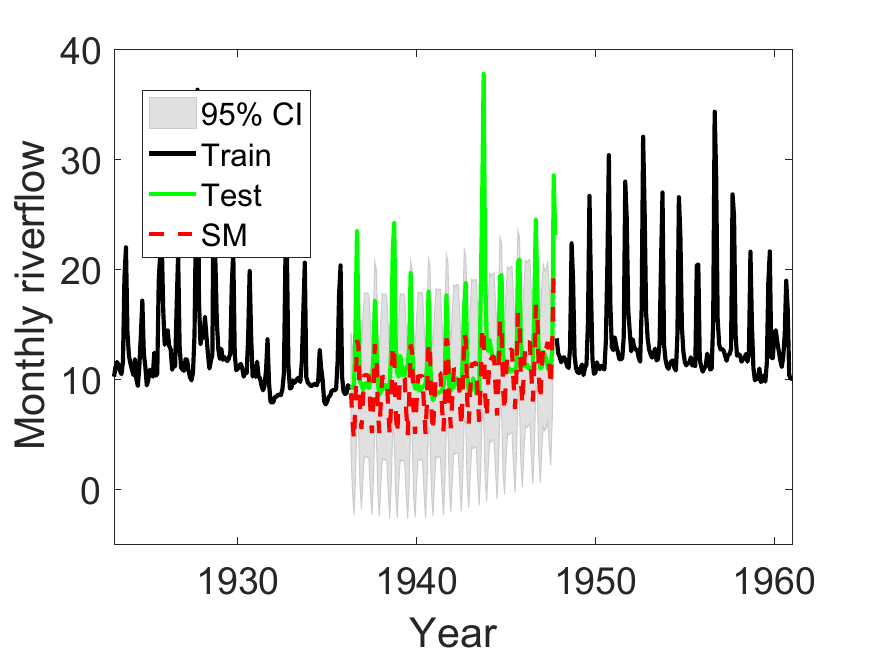}
            & \figart{0.45}{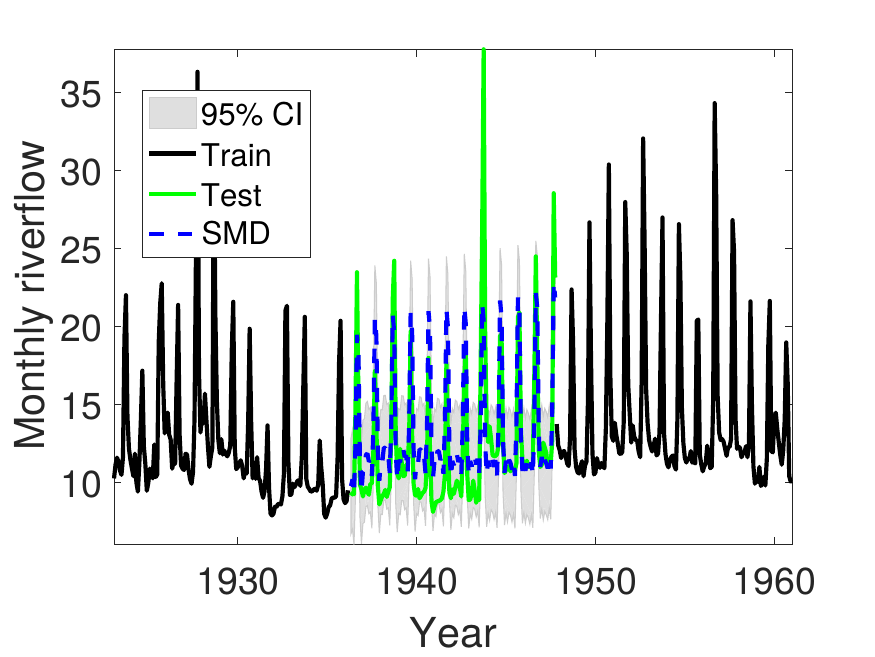}
        \end{tabular}
        \caption{Interpolations of GPs with the SM (left) and SMD (right) kernels on the monthly river flow dataset.}
        \label{fig:riverflow}
    \end{figure}

    \tbf{Posterior dependency between SM components:}
    On the other hand, by computing the posterior covariance between two functions, conditioned on their sum \cite{Duvenaud2014}, we obtain posterior cross covariance between two components as, $\operatorname{Cov}\left({f}_{\sm, i}^{*}, {f}_{\sm, j}^{*}|{\vf}_{\sm, i+j}\right)=-{\vk_{\sm, i}^{*\top}}{K_{\sm, i+j}^{-1}}{\vk_{\sm, j}^{*}}$, 
    where ${\vf}_{\sm, i+j}={\vf}_{\sm, i}+{\vf}_{\sm, j}$ and
    $K_{\sm, i+j}=K_{\sm, i}+K_{\sm, j}$.
    As investigated in \cite{Duvenaud2014}, the posterior cross covariance can indicate the underlying dependency between $f_{\sm, i}$ and $f_{\sm, j}$.
    We further normalize the posterior cross covariance as posterior correlation coefficient $\rho_{ij}$ with range $[-1, 1]$:
    $\rho_{ij}=\frac{\operatorname{Cov}\left({f}_{\sm, i}^{*}, {f}_{\sm, j}^{*}|{\vf}_{\sm, i+j}\right)}{\big(\bbV({f}_{\sm, i}^{*}|{\vf}_{\sm, i+j})\bbV({f}_{\sm, j}^{*}|{\vf}_{\sm, i+j})\big)^{1/2}}$.
    Note that there is no underlying dependency if $\rho_{ij}=0$, otherwise, $f_{\sm, i}$ and $f_{\sm, j}$ are dependent. 
    In \fref{covar_riverflow}, the left subplot shows high and complex posterior correlation coefficient $\rho_{56}$ of the SM. 
    In the right subplot, the intensity of the dependency structure in the SMD is indicated by the positive and negative values of $\gamma_{14}$.
    Note that there is no alignment between SM and SMD components because they are separately optimized.
    
    \begin{figure}[h!]
        \scriptsize
        \renewcommand{\tabcolsep}{2.0mm}
        \def\figcovar#1#2{\includegraphics[width=#1\columnwidth]{fig-GCSM-SM-Q10-riverflow_Madison-Covar-#2}}
        \def\figgamma#1#2{\includegraphics[width=#1\columnwidth]{fig-GCSM-SMDtp-Q10-riverflow_Madison-Gamma-#2}}   
        \begin{tabular}{p{1mm}*{2}{c}}
            & \figcovar{0.25}{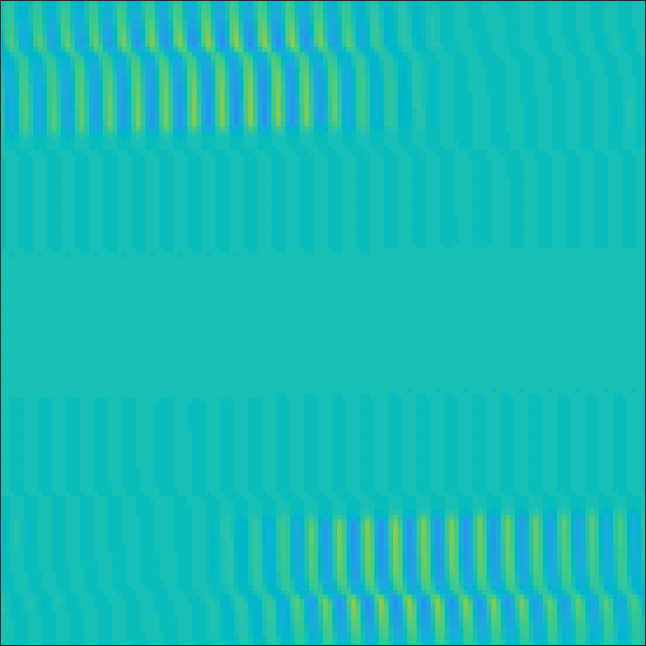} & \figgamma{0.25}{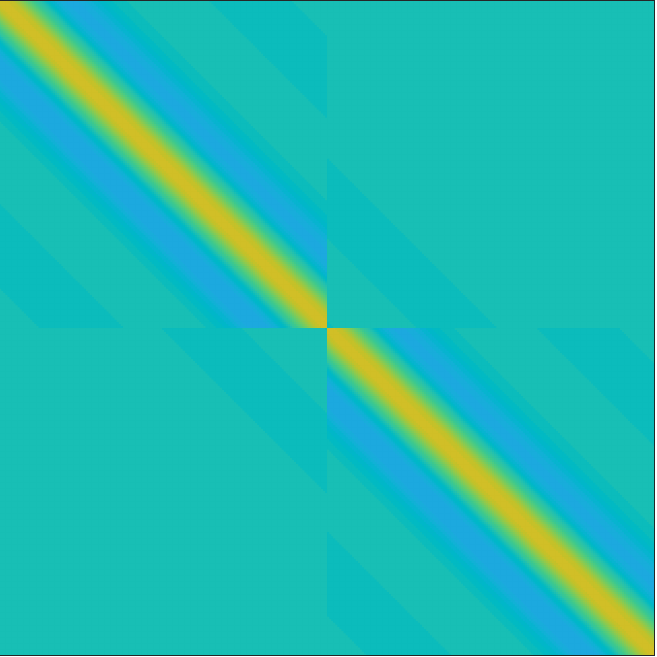} \\ 
        \end{tabular}
        \centering
        \begin{tabular}{p{2mm}*{1}{c}}
            & {} \\
            \includegraphics[width=8mm, clip, trim=12.0cm 0.8cm 0.0cm 0.5cm]{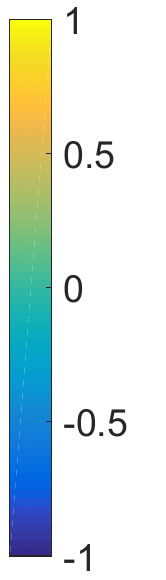} 
        \end{tabular}
        \caption{The posterior correlations $\rho_{ij}$ (left) and the   dependency structure intensity $\gamma_{ij}$ (right) for the SM and SMD kernels on river flow, respectively.}
        \label{fig:covar_riverflow}
    \end{figure}

    \subsection{Dependency structure for joint interpolation and extrapolation of yearly sunspots modeling}\label{sec:sunspot}
    In addition to the interpolation task, we simultaneously perform interpolation and extrapolation to further substantiate the learning ability of the SMD. We consider the yearly sunspot number dataset\footnote{http://www.sidc.be/silso/infossntotyearly}  collected between  1700 and 2014. The historical evolution of yearly sunspots can help explain spatial magnetic field environment changes affected by sun activities. The yearly sunspot number is obtained by taking an arithmetic mean of the daily total sunspot number over all days of each year. Sunspots appear darker than the surrounding areas on the sun's photosphere \cite{WILLSON700}. They usually have lower surface temperatures than the areas around them. A sunspot has an irregular period of existence: the average number of sunspots that are monitored increases and decreases with a quasi-period. There are dependencies between patterns of sunspots caused by some physical types of interference. 
    As shown in \fref{sunspot}, patterns in yearly sunspots contain various irregular peaks over 315 years.

    There are 315 records in the dataset. We use the last $10\%$ of the data for the extrapolation test (in solid green) and randomly sample $20\%$ of the original data from the first $90\%$ of the yearly sunspots as the interpolation test (in crossed green). The remaining $70\%$ of the data are used for training (in black). The legends Test$_{ext}$ and Test$_{int}$ denote the extrapolation and interpolation testing data, respectively. 
    Note that the training data are not equally sampled due to missing values. We initially considered $Q=20$ components for both SMD and SM. Time and phase delays in the SMD are also initialized as zeros.

    \begin{figure}[h!]
        \centering
        \renewcommand{\tabcolsep}{1.0mm}
        \def\figart#1#2{\includegraphics[width=#1\columnwidth, clip, trim=0.1cm 0.0cm 0.1cm 0.1cm]{fig-GCSM-#2-Q10-yearly-sunspot-Full.pdf}}
        \begin{tabular}{p{0.5mm}*{2}{c}}
            & \figart{0.4}{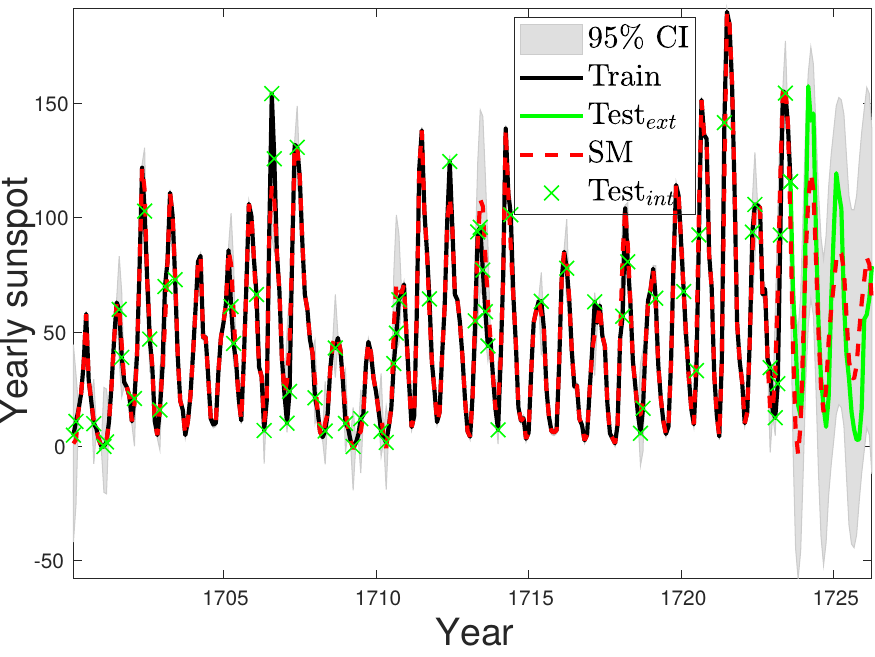}
            & \figart{0.4}{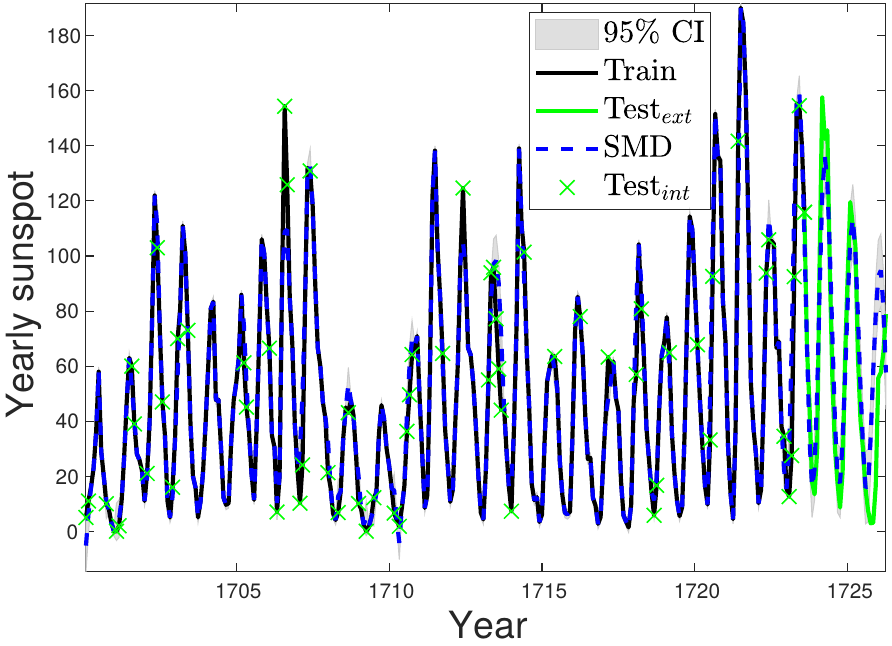}\\
        \end{tabular}
        \caption{Interpolations and extrapolations of GPs with the SM (left) and SMD (right) kernels on the yearly sunspot dataset.
        }
        \label{fig:sunspot}
    \end{figure}
    
    
    \begin{figure}[h!]
        \scriptsize
        \renewcommand{\tabcolsep}{0.5mm}
        \def\figcrossspec#1#2{\includegraphics[width=#1\columnwidth]{fig-GCSM-SMDtp-Q10-yearly-sunspot-SpecDep-#2}}
        \def\figcrosscov#1#2{\includegraphics[width=#1\columnwidth]{fig-GCSM-SMDtp-Q10-yearly-sunspot-CovDep-#2}}
        \centering
        \begin{tabular}{p{1mm}*{2}{c}}    
            & \figcrossspec{0.4}{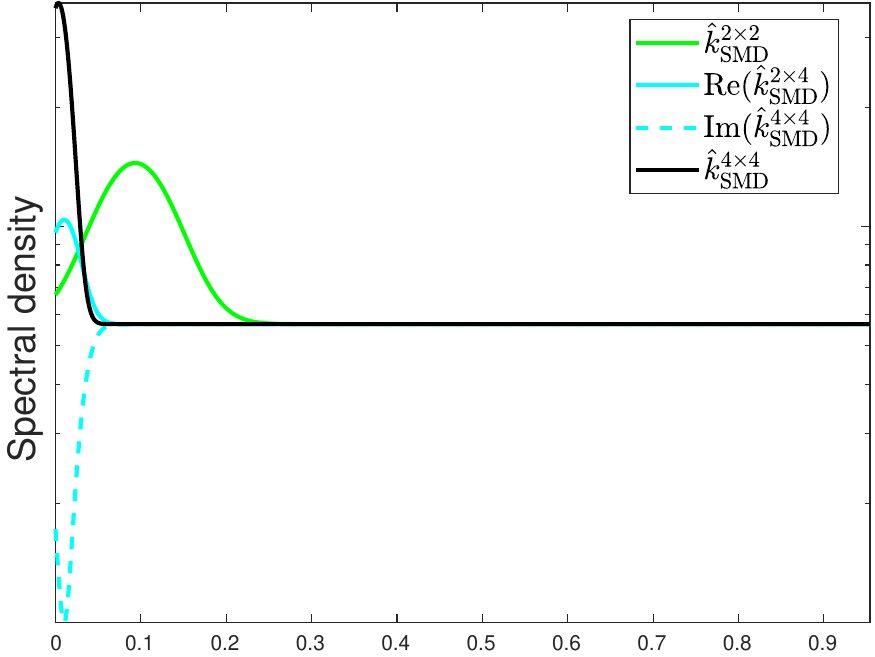} & \figcrosscov{0.4}{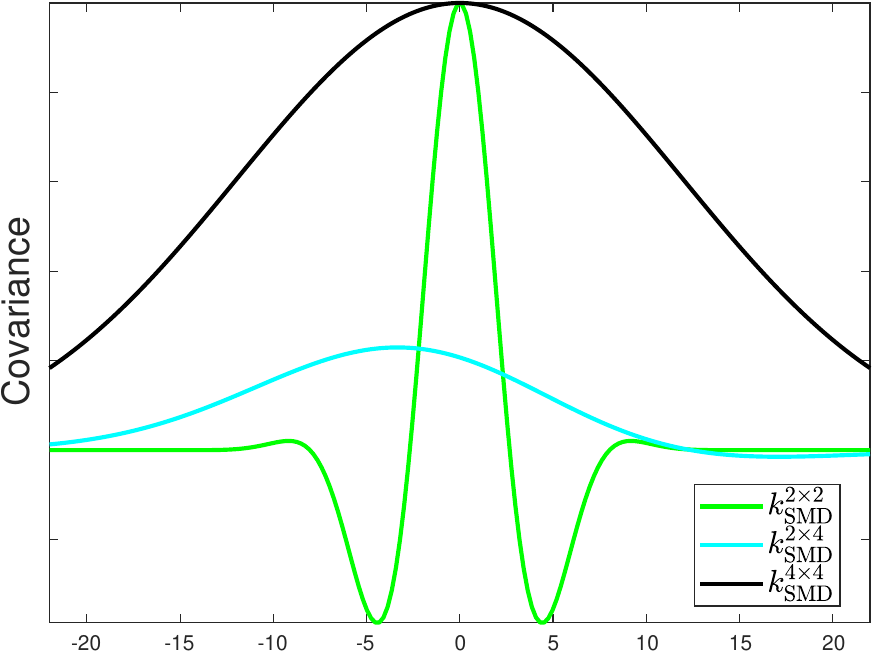}  \\ 
        \end{tabular}
        \caption{The learned dependency structures on the yearly sunspot dataset. 
            Left subplot: the SDs of $\freq{k}_{\gcsm}^{2\times{2}}$, real part  $\mathrm{Re}(\freq{k}_{\gcsm}^{2\times{4}})$, imaginary part  $\mathrm{Im}(\freq{k}_{\gcsm}^{2\times{4}})$, and $\freq{k}_{\gcsm}^{4\times{4}}$ are in green, solid cyan, dashed cyan, and black, respectively. Right subplot: the covariances of ${k}_{\gcsm}^{2\times{2}}$, ${k}_{\gcsm}^{2\times{4}}$, and ${k}_{\gcsm}^{4\times{4}}$.
        }
        \label{fig:covar_sunspot}
    \end{figure}

    As shown in subplots (a) and (b) of \fref{covar_sunspot}, there is a significant TP delay dependency structure (in cyan) between components $2$ (in green) and $4$ (in black) in the SMD. As investigated in \sref{sparse-dep}, $\freq{k}_{\gcsm}^{2\times{4}}$ is significant because component 4 and component 2 are close to each other.
    Due to the TP delays, the covariance of ${k}_{\gcsm}^{2\times{4}}$ is shifted to left. 
    The period of the dependency structure ${k}_{\gcsm}^{2\times{4}}$ is smaller than the $2$nd component and larger than the $4$th component and has a connection to the location of $\freq{k}_{\gcsm}^{2\times{4}}$. 
    From the CR and SR shown in \tref{CR}, the SMD using the SA algorithm can reduce the hyperparameter size by 40.7\% and dependency structures by 50.3\%.
    The above results indicate that both the SMD and SM can interpolate missing values well with a small CI. However, for the extrapolation task, the SMD achieves better performance and confidence (see \tref{MSE} and \fref{sunspot}).  
    With this experimental result, 
    we can conclude that the SMD can perform interpolation and extrapolation equally well for incomplete signals.
    %
    \subsection{Scalable SMD on large scale multidimensional data}\label{sec:scalable}

    Furthermore, we comparatively evaluate the scalable SMD on a large multidimensional abalone\footnote{http://archive.ics.uci.edu/ml/datasets/abalone} dataset. 
    We apply automatic relevance determination (ARD) for other baseline kernels to remove irrelevant input. Note that the FBM, ULL, and NSM kernels are not applicable to multidimensional datasets ($P>2$).
    When modeling large data, exact inference \cite{Wilson2014a,Williams2001,Quinonero-Candela2005,Snelson2006} of GP is prohibitively expensive and meets $\compO(n^{3})$ computational complexity  and $\compO(n^{2})$ memory. The 
    expensive computation cost involves computing the inverse and determinant of $K+\sigma^{2}_{n}I$.
    We consider a scalable SMD using stochastic variational inference (SVI) framework \cite{Hensman2013,Hensman2017}.
    Specifically, SVI can approximate the underlying GP posterior with a GP conditioned on a small set $U$ with $m$ inducing points. 
    The inducing points $U$ can be seen as a set of global variables summarizing the structure of the large training data to perform variational inference. The variational distribution $p(\vu)=\N(\vu; \vmu_{\vu}, \Sigma_{\vu})$ with
    \begin{align}
        \begin{split}
            \Sigma_{\vu} =&K_{UU}^{-1} + \sigma^{-2}_{n}K_{UU}^{-1}K_{U{X}}K_{U{X}}^{\top}K_{UU}^{-1},\\
            \vmu_{\vu} =&\sigma^{-2}_{n} \Sigma_{\vu}^{-1}K_{UU}^{-1}K_{U{X}}{\vy},
        \end{split}
    \end{align}
    gives a variational lower bound ${\Low_{3}(\vu; \vmu_{\vu}, \Sigma_{\vu})}$, also called evidence lower bound (ELBO) of the quantity $p(\vy|X)$, satisfying $\log{p(\vy|X)}\geq{\Low_{3}(\vu; \vmu_{\vu}, \Sigma_{\vu})}$, 
    where $K_{U{X}}=k_{\gcsm}(U, X)$
    and $K_{UU}=k_{\gcsm}(U, U)$.
    As mentioned in \cite{Hensman2013}, the variational distribution $p(\vu)$ contains all the information encoded into the posterior approximation, which describes the distribution of function values at the inducing points $U$. Letting $\frac{\partial{\Low_{3}}}{\partial\vmu_{\vu}}=0$ and $\frac{\partial{\Low_{3}}}{\partial\Sigma_{\vu}}=0$, we can approximate the optimal solution of the variational distribution. 
    Therefore, we have the posterior distribution of a test point as $p(y^{*} | \vx^{*}, X, Y)= \N(\tilde{y}^{*}, \bbV[y^{*}])$,
    where the predictive mean $\tilde{y}^{*}=\vk^{*}_{U} K_{UU}^{-1}\vmu_{\vu}$, predictive variance $\bbV[y^{*}]=k^{**}+{\vk^{*\top}_{U}}(K_{UU}^{-1}\Sigma_{\vu}K_{UU}^{-1}-K_{UU}^{-1})\vk^{*}_{U}$, $k^{**}=k_{\gcsm}(\vx^{*}, \vx^{*})$, and $\vk^{*}_{U}=k_{\gcsm}(U, \vx^{*})$.
    Finally, the complexity of the SMD on a large dataset is reduced to $\compO(m^{3})$.

    The abalone\footnote{http://archive.ics.uci.edu/ml/datasets/abalone} dataset has
    4177 samples with eight attributes: sex, length, diameter, height, whole weight, shucked weight, visceral weight, and shell weight. We aim to predict the age of abalone, which is usually measured by physical assessment. Generally, an abalone's age is measured by cutting the shell through the cone, staining it, and counting the number of rings through a microscope. The number of rings directly reflects the age of an abalone. The task is to predict the number of rings from the eight attributes.  
    Specifically, we use the first 3377 instances as training data and the remaining 800 instances as testing data. We use $Q = 10$ components for the SMD and SM due to the explosive expansion of hyperparameter space for multidimensional input. Note that we set the number of inducing points $m=500$ in the SVI. The SMD and SM use the same hyperparameters initialization described in the SA algorithm. 
    The results in \tref{MSE}
    show that on this type of task, the SMD also performs better, with a lower MSE than the SM. 
    
    \subsection{Discussion}
    
    
    \begin{table}[!ht]
        \scriptsize
        \caption{Performances of various baselines versus the SMD in terms of MSE. RI denotes random initialization (RI).
        }
        \label{tab:MSE}
        \begin{center}
            \renewcommand{\tabcolsep}{1.5mm}
            \newcolumntype{C}{>{\centering\arraybackslash}m{3.5cm}}
            \begin{tabular}{C r @{.} l r @{.} l  r @{.} l r @{.} l r @{.} l r @{.} l r @{.} l r @{.} l 
                    r @{.} l 
                }
                \toprule
                {Kernel} &
                \multicolumn{2}{c}{Synthetic} &
                \multicolumn{2}{c}{Riverflow} &
                \multicolumn{2}{c}{Sunspot} &		
                \multicolumn{2}{c}{Abalone} 
                \\
                \midrule
                
                LIN  & {0}&{32\std{0.11}} & {24}&{37\std{4.57}} & {1617}&{45\std{131.63}} 
                & {10}&{93\std{2.07}}
                \\
                
                SE  & {0}&{31\std{0.20}} & {174}&{14\std{19.23}}  & {591}&{24\std{9.21}} 
                & {8}&{14\std{3.24}}  
                \\
                
                Poly  & {0}&{32\std{0.18}} & {182}&{01\std{18.73}} & {1627}&{02\std{127.84}}   
                & {6}&{30\std{1.71}}
                \\
                
                PER  & {0}&{35\std{0.13}} & {20}&{31\std{7.22}} & {1533}&{84\std{143.15}} 
                & {7}&{98\std{2.80}} 
                \\
                
                RQ  & {0}&{31\std{0.10}} & {24}&{88\std{6.32}} & {307}&{22\std{30.38}} 
                & {5}&{38\std{1.53}}  
                \\
                
                MA  & {0}&{32\std{0.14}} & {170}&{29\std{18.90}} & {590}&{72\std{26.21}} 
                & {7}&{52\std{1.05}}  
                \\
                
                Gabor & {0}&{31\std{0.20}} & {22}&{51\std{3.37}}  & {4047}&{83\std{19.78}} 
                & {3}&{68\std{1.13}} 
                \\
                
                FBM  & {0}&{49\std{0.19}} & {23}&{84\std{3.21}} & {6428}&{69\std{515.98}} 
                & {-}&{-}  
                \\
                
                ULL & {0}&{26\std{0.11}} & {168}&{06\std{18.16}} & {467}&{08\std{23.90}}  
                & {-}&{-}  
                \\
                
                NN  & {0}&{32\std{0.16}} & {23}&{65\std{2.32}} & {1522}&{67\std{68.58}} 
                & {3}&{61\std{1.07}}  
                \\
                
                NSM  & {0}&{53\std{0.21}}  & {166}&{85\std{20.45}} & {4697}&{24\std{375.15}} 
                & {-}&{-}  
                \\
                
                SM (RI)  & {0}&{41\std{0.16}}  & {23}&{55\std{2.79}} & {363}&{64\std{64.32}} 
                & {3}&{64\std{1.12}}  
                \\
                
                SM (BHI) & {0}&{43\std{0.18}}  & {13}&{91\std{1.78}} & {270}&{78\std{13.96}}
                & {3}&{52\std{0.51}} 
                \\
                
                {SMD (RI, ${\vtime=\vzero, \vphase=\vzero}$)}  & {0}&{47\std{0.13}}  & {23}&{31\std{4.76}} & {329}&{81\std{18.19}}
                & {3}&{56\std{1.90}} 
                \\                
                
                SMD (BHI, ${\vtime=\vzero, \vphase=\vzero}$)  & {0}&{28\std{0.14}}  & {9}&{55\std{1.79}} & {184}&{47\std{18.58}}
                & {3}&{37\std{0.42}} 
                \\                
                
                SMD (RI, ${\vtime\neq\vzero, \vphase\neq\vzero}$) & {0}&{45\std{0.95}}  & {19}&{26\std{3.53}} & {322}&{50\std{21.67}}
                & {3}&{74\std{1.31}} 
                \\
                
                SMD (BHI, ${\vtime\neq\vzero, \vphase\neq\vzero}$) & \tbf{0}&\tbf{05\std{0.01}} & \tbf{8}&\tbf{96\std{0.72}} & \tbf{171}&\tbf{16\std{13.10}} 
                &  \tbf{3}&\tbf{25\std{0.38}} 
                \\
                \bottomrule
            \end{tabular}
        \end{center}
    \end{table}
    
    
    \begin{table}[!ht]
        \scriptsize
        \caption{The average CRs and SRs of the SM and SMD variants using SA.}
        \label{tab:CR}
        \begin{center}
            \newcolumntype{C}{ >{\centering\arraybackslash}m{4.0cm}}
            \newcolumntype{c}{ >{\centering\arraybackslash}m{2.0cm}}
            \renewcommand{\tabcolsep}{-0.2mm}
            \begin{tabular}{C c c c c}
                \toprule
                {Kernel} & {CR Riverflow} & {CR Sunspot} 
                & {SR Riverflow} & {SR Sunspot}
                \\
                \midrule                             
                SM &{33.2\%} &{30.3\%} 
                &{--} &{--} 
                \\                      
                
                SMD (BHI, ${\vtime=\vzero, \vphase=\vzero}$) &{35.3\%} &{32.8\%} 
                &{89.1\%} &\tbf{57.6\%} 
                \\
                
                SMD (BHI, ${\vtime\neq\vzero, \vphase\neq\vzero}$) &\tbf{38.9\%} &\tbf{40.7\%}
                &\tbf{89.3\%} &{55.3\%} 
                \\
                \bottomrule
            \end{tabular}
        \end{center}
    \end{table}

    In \tref{MSE}, the results demonstrate that the SMD ($\vtime\neq\vzero, \vphase\neq\vzero$) performs better than other baselines as well as the other SMD variants ($\vtime=\vzero$ or $\vphase=\vzero$). 
    Our experiment analysis indicates that signals containing dependency structure caused by physical interference can be learned well by a GP with SMD kernel.    
    The SM and SMDs using BHI perform much better than those using random initialization (RI). The proposed BHI in the SA algorithm can provide a good starting point for optimization in high-dimensional hyperparameter space.

    As shown in \tref{CR}, the CRs of both the SM and SMD are larger than $30\%$ due to the use of the SA algorithm. Hence, the SA algorithm can much reduce the hyperparameter space of SMD and SM.
    In \tref{CR}, the SMDs with SA usually have better CR than the SM. This may be caused by the fact that the SMD with dependency structures has a better representation ability than the SM. The SMD can describe an underlying function with fewer components than the SM because the latter needs additional components to delineate the latent dependency structure. In addition, \tref{CR} shows that the SR of all the SMDs is high, which means that the dependency structures are sparse. Most dependency structures are tiny and removable. 
    Due to the higher number of hyperparameters (at least three times than SM) and overfitting troubles, the results in \tref{MSE} show the unsatisfactory performance of NSM on the synthetic signal and on real-world datasets.
    
    
    \section{Conclusion}\label{sec:conclude}
    We propose a novel SMD kernel, which extends the SM kernel by incorporating TP delayed dependency structures.
    An interpretable SA algorithm for the SMD is introduced to effectively initialize its hyperparameters, compress components, and obtain sparse dependency structures automatically.

    The results of extensive experiments on both the synthetic and real-life datasets indicate that the SMD using the structure adaptation (SA) algorithm can learn TP delayed dependency structures between the components and perform more accurate interpolation and extrapolation. Hence, the benefits of SMD are shown to be significant.

    Two main issues remain to be addressed in future work. The first issue is the initialization of the TP parameters. Here, we simply initialized them as zeros. However, more tailored, effective methods remain to be investigated. Another issue, common to all GP methods, is the problem of sparse or efficient inference \cite{Rasmussen2006,Quinonero-Candela2005}, which also needs to be further improved for GPs with the SMD on big data.
    
    \section*{Acknowledgment}
    We would like to thank Elena Marchiori, Twan van Laarhoven, and Perry Groot for their comments on a past version of this work. 
    This research is supported by the Guangdong Provincial Key Laboratory of Future Networks of Intelligence, The Chinese University of Hong Kong, Shenzhen, under Grant No. 2022B1212010001. The work was partly supported by the Natural Science Foundation of China (NSFC) with grant No. 62106212, 
    by the Natural Science Foundation of Hunan Province, China, under Grant 2023JJ40689, and by the High Performance Computing Center of Central South University (CSU). The work of Feng Yin was supported by the NSFC with Grant No. 62271433.

\bibliographystyle{elsarticle-num}
\bibliography{main.bbl}


\end{document}